# Semantic Segmentation based Scene Understanding in Autonomous Vehicles

Thesis presented by

EHSAN RASSEKH

Master of Artificial intelligence

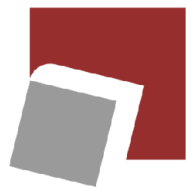

Institute for Advanced Studies
in Basic Sciences
Gava Zang, Zanjan, Iran

# Abstract

In recent years, the concept of artificial intelligence (AI) has become a prominent keyword because it is promising in solving complex tasks. The need for human expertise in specific areas may no longer be needed because machines have achieved successful results using artificial intelligence and can make the right decisions in critical situations. This process is possible with the help of deep learning (DL), one of the most popular artificial intelligence technologies. One of the areas in which the use of DL is used is in the development of self-driving cars, which is very effective and important. In this work, we propose several efficient models to investigate scene understanding through semantic segmentation. We use the BDD100k dataset to investigate these models. Another contribution of this work is the usage of several Backbones as encoders for models. The obtained results show that choosing the appropriate backbone has a great effect on the performance of the model for semantic segmentation. Better performance in semantic segmentation allows us to understand better the scene and the environment around the agent. In the end, we analyze and evaluate the proposed models in terms of accuracy, mean IoU, and loss function, and the results show that these metrics are improved.

# TABLE OF CONTENTS

# List of Tables

# LIST OF FIGURES

INTRODUCTION AND MOTIVATION

## 1.1 Introduction

The safety of the driver and occupants of the car has always been the best motivation and goal for progress in the automotive industry. Hence, carmakers are constantly thinking about new developments and innovations. According to available reports, the average annual number of deaths and injuries in traffic accidents worldwide is increasing by 1.3 million deaths and between 20 and 50 million injuries.[1]

According to [13], more than 90% of accidents are caused by human error. Therefore, the activity in the field of self-driving cars has increased in recent years, and due to the advances that have been made in the field of artificial intelligence, are expected to be minimized these human errors. In recent years, big technology companies such as (Google, Tesla, Apple, Nvidia, Uber, Lift, etc.) as giants in the field of artificial intelligence, have paid special attention to the issue of driverless cars. Self-driving vehicles can be classified into following five levels in terms of capability and intelligence. This classification is based on the action and reaction that the vehicles show in the environment[2][14].

The Full Autonomous Driving Process can be divided into three important steps as shown in Fig. 1.2. Each part contains different tasks. Any small breakdown can have many negative effects on the final behavior of the vehicle. The part of perception is the part where innovative and new methods of artificial intelligence can help a lot.

---

[1]https://www.asirt.org/safe-travel/road-safety-facts/
[2]https://www.iotforall.com/5-autonomous-driving-levels-explained

| Levels | Description |
|--------|-------------|
| Level Zero – No Automation | At level zero, all the work of driving the car is done by the driver. |
| Level One – Driver Assistance | At this level, the car can only assist and optimize certain operations. However, the driver is still responsible for accelerating, braking and monitoring the environment. |
| Level Two – Partial Automation | Recently, most automakers have focused on this level. Vehicles using sensors and cameras can help the driver control acceleration and steering. However, the driver still must prepare himself and control his environment so that there is no safety problem in case of an emergency. |
| Level Three – Conditional Automation | At level 3, the car can monitor and control its environment using sensors such as the LIDAR. Most vehicles at this level do not need driver control and attention for speeds of less than 37 miles per hour. In 2018, Audi announced that it will produce cars at this level. |
| Level Four – High Automation | At level 4, the car can monitor steering, braking, acceleration, rerouting, etc, which is done by signals received from sensors. |
| Level Five – Complete Automation | At this level of autonomous driving, the car does not need human supervision at all. All important operations and monitoring the environment and identifying the environment is done by the vehicle. Also, the vehicle can control the traffic. NVIDIA has introduced an artificial intelligence computer to help with this level of autonomous driving. |

Figure 1.1: Different levels of Autonomous Vehicles

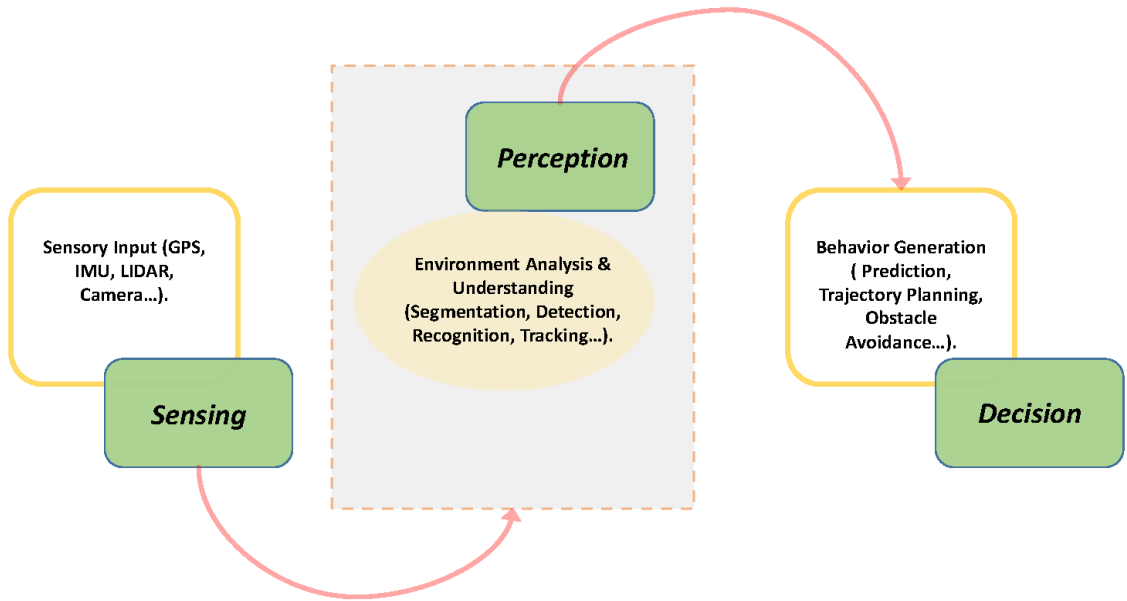

Figure 1.2: The flowchart of the full autonomous driving process (Level 5)

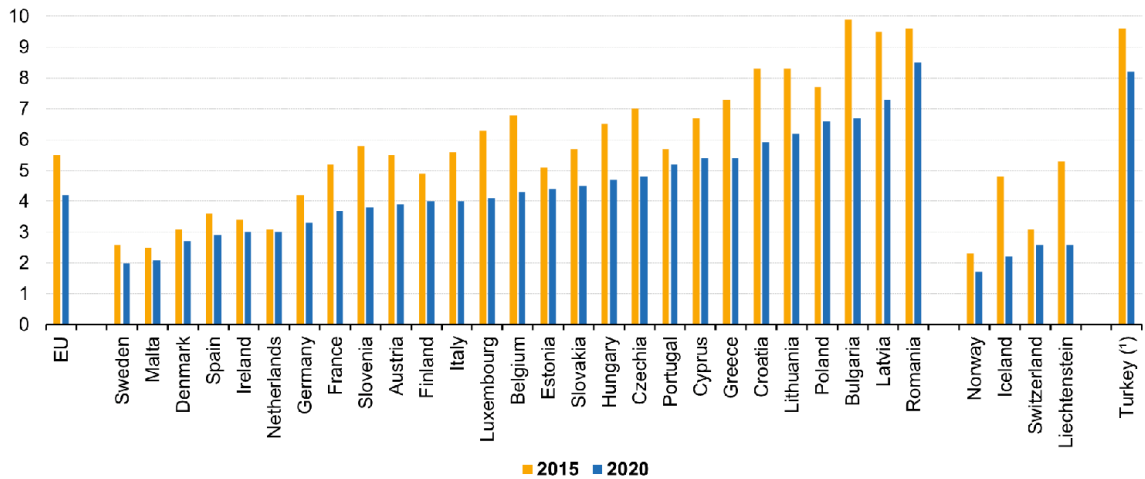

**Road traffic deaths, by country, 2015 and 2020**
(number per 100 000 people)

(¹) 2018 data (instead of 2020).
*Source:* European Commission services, DG Mobility and Transport (Eurostat online data code: sdg_11_40)

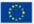

Figure 1.3: Road traffic deaths, by country, 2015 and 2020 (number per 100000 people)

## 1.2 Motivation and Objectives

On May 7, 2016, a middle-aged man named Joshua Brown was killed when his Tesla Model S sedan collided with a trailer. Now, three years later, another person has lost his life in a similar way. Jeremy Byrne, 50 years old, the owner of a Tesla Model 3 car, collided again with a tractor-trailer passing by and cut the roof. The similarity between the two accidents is that both drivers use Teslas advanced driver assist system Autopilot in the event of an accident. Tesla's autopilot system is a Level 2 semi-automatic system with features such as cruise control, lane-keeping assistant, automatic parking, and automatic lane change. Tesla claims that it is one of the safest autopilot systems in the world today, but the deaths of Brown and Banner cast doubt on that claim. The above two incidents are different from each other. For example, the car of these two people, in terms of driving assistance technology, was very different from each other. Of course, both systems are called autopilots. But Mr. Brown's autopilot system used Mobileye technology. Mr. Banner's car was a Tesla Model 3, equipped with a second-generation autopilot system. A system that Tesla has produced and developed exclusively.

After Mr. Brown's death, Tesla announced that the car's camera could not detect the white trailer in the bright surroundings. In addition, the occupant himself did not pay attention to the road, which caused this to happen. Vehicle safety experts say cruise control systems, such as autopilot systems, use radar to prevent accidents. Radars are effective at detecting moving objects, but they have difficulty detecting stationary objects. In addition, the trailers were horizontally inside the road in both accidents, which was not in line with the direction of Tesla's vehicle. Because of this, autopilot systems radar could not detect the obstacle. However, various data from the camera must be trained in the above systems. Such as cars or obstacles that are located or moving in different directions relative to a person's car. According to experts, machine learning technologies and artificial intelligence have limitations. If the sensors see something they have not seen before, they will be able to detect a low percentage.

Using a camera sensor, a computer creates images by calculating pixel colors from grids of numbers. For example, converting these numbers into coordinates of objects in the image is far from obvious. A machine learning algorithm can be used to learn such a complicated mapping function from input-output pairs instead of encoding it by hand. A growing field of research known as computer vision involves creating high-level representations of the scene with the aid of cameras as perceptions. By recognizing objects and textures in the scene, these models capture geometric information via structure-from-motion techniques, and semantic information via object recognition. In recent years, deep learning has emerged as a new paradigm for machine learning, enabling new levels of performance on tasks ranging from natural language processing to computer vision. Feedforward neural networks are composed of simple parameterised functions called layers and are used to structure large parameterised functions. A training process is used to learn the parameters of these networks by observing labeled data. It's common for computer vision recognition problems to use a specific topology of feedforward neural networks, known as Convolutional Neural Networks (CNNs). In addition to recognizing objects, deep learning has been applied to more demanding computer vision tasks, including semantic segmentation, and scene understanding, the focus of this thesis. According to the semantic segmentation problem, each pixel in an image is assigned a label, such as "road", "sky", "cars", or "pedestrian". It is essential that the data provide sufficient examples to cover all lighting conditions and adverse weather conditions in order to perform image-based semantic segmentation of driving scenes. The visual data collected by camera sensors will remain an important component of autonomous driving regardless of what the future holds. Vehicle cameras record the surrounding environment, such as urban scenes or highways. Several factors contribute to making a driving decision: lane markings, traffic signs, other vehicles, trucks, bicyclists, a group of pedestrians at a crosswalk, and obstacles on the road. Visual scene understanding algorithms have to be capable of detecting individual objects, classifying them, and delineating their boundaries tightly in order to assess their depth and shape. Parsing the scene into image regions is necessary to determine whether the scene is composed of roads, pavement, trees, and buildings.

In the field of self-driving cars, the ultimate goal of automakers is to achieve a system that can achieve artificial performance and even better than a human performance by using artificial intelligence and deep learning models. Achieving such a goal requires that the vehicle must have a completely accurate and reliable understanding of its environment in order to be able to quickly analyze any situation and make the right decision. The technology of visual scene understanding has a wide range of applications. In this thesis, most of the experiments focus on autonomous vehicle datasets, but the methods can also be applied to other domains. As long as a labeled dataset is available for the specific domain, deep learning and computer vision techniques can be used to re-train the same models. In each of these applications, scene understanding algorithms must be capable of detecting, classifying, and segmenting objects and image regions, and estimating object parameters, as well as predicting short-term behavior.

## 1.3 Project Goals

**Research Question:** How effective can the Semantic segmentation be in improving agent interaction with the environment and scene understanding? Will the use of different backbones affect improving the scene understanding? How can Deep Learning (DL) help us to design semantic segmentation models?

In this work, we focus on improving the Semantic Segmentation based Scene Understanding. We use the BDD100k [12] dataset to conduct our research. Our goal is to improve the semantic segmentation task for Scene Understanding. We aim to find a better architecture for semantic segmentation. In this regard, we propose several efficient models to investigate scene understanding through semantic segmentation. This thesis makes contributions to the field of scene understanding, using deep learning to perform semantic segmentation in computer vision. We propose a novel compound model for semantic segmentation datasets to evaluate the BDD100k [12] dataset of urban driving scenes. This project's source code repository is publicly available at the following URL: https://github.com/EhsanR47

# 1.4 Thesis Structure

**Chapter 1** *Introduction* - In this chapter, we first looked at the statistics of deaths in road accidents. Then we discussed the companies that are active in this field. Next, we examined different levels of self-driving vehicles. To explore the problems in this field, we mentioned two examples of Tesla's accidents. Finally, we discussed our goals and motivation for doing this work.

**Chapter 2** *Related work and Background information* - In this chapter, the theoretical background for the content of this thesis is presented. First, we investigate the background of the knowledge graph in the field of scene understanding and then examine its challenges. Then, we discuss the background in the fields of semantic segmentation with a specific emphasis on deep learning techniques.

**Chapter 3** *Review of Datasets for Autonomous Vehicles* - This chapter we reviewed important datasets in the field of Scene Understanding. We discussed about details of them. Finally, we compared them.

**Chapter 4** *Scene Understanding Methods and Implementation Details* - We discuss the methods used and implementation details. We investigate the neural network structure of each model in detail. Next, we talk about transfer learning. Finally, we describe how to prepare the data set.

**Chapter 5** *Experimental Results* - This chapter presents the results of the methods described in the previous chapter. We want to discuss and analyze our experimental results for the scene understanding methods and the semantic segmentation models. First, we will discuss the dataset BDD100K and then evaluate each model with different approaches in terms of the loss function, accuracy, and mean Intersection over Union (mIoU).

**Chapter 6** *Conclusion and Future Work* - Summarise and Conclusion the results observed in this thesis and provides suggestions for further work.

RELATED WORKS AND BACKGROUND INFORMATION

## 2.1  Knowledge Graph based Scene Understanding

In May 2012, Google introduced the first version and interpretation of a Knowledge Graph (KG), although the term "knowledge graph" has been used in old times. Afterwards, KG has been used in other large companies such as Facebook, LinkedIn, Amazon, Microsoft, IBM and Uber, etc. for a variety of purposes, including better information structuring as well as better interaction with users [15]. Recently, KG and graph structure are used as a database in various industries, such as: banking, automotive industry (for better performance of self-driving vehicles), pharmacy and health, media and etc. The KG is a semantic network consisting a set of entities (which includes properties) and the relationship that exists between them. Which can be expressed as G = (E, R) Showed [15]. The following figure shows a simple KG [16]. Which includes nodes and the relationships between them and each node has its own properties [1].

Recently, with the development of self-driving vehicles, the use of KG has become especially important. Since the amount of information received by sensors and cameras in a scene of driving is very large and this information is constantly updated, to use artificial intelligence and machine learning methods by which machine learning models can be taught. We need the received data to be properly structured so that we can better manage and organize them[14].Using semantic knowledge to organize and give meaning to the objects in the scenes, which constitute our data, can improve performance and speed up learning, and then perform and select the appropriate action to deal with factors in driving, such as pedestrian, dealing with obstacles, traffic signs, choosing a better and shorter route, as well as reducing traffic, etc. are very effective.Using semantic knowledge and then structuring the data obtained by the KG can help our agent (car) to perform better in an Reinforcement learning environment and perform better actions according to the knowledge it acquires[17][18].

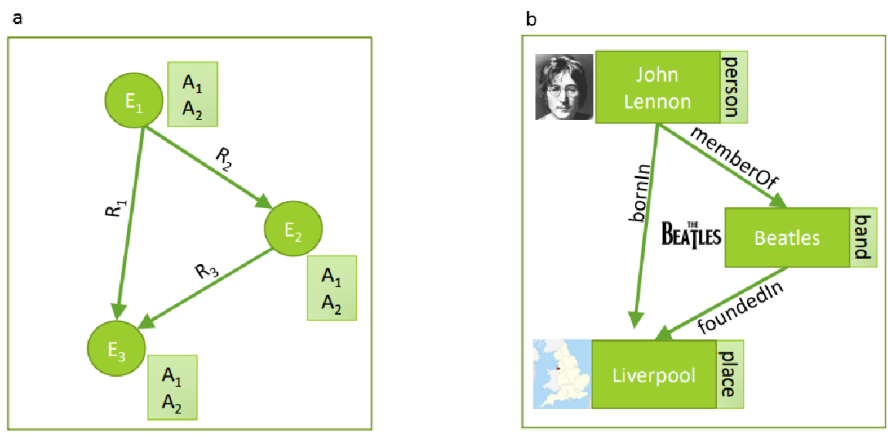

Figure 2.1: Example of a Knowledge Graph [1]

According to [17], they focus on meanings using semantic priors in their work to navigate. They also used Graph Convolutional Networks to integrate prior knowledge into a deep Reinforcement learning (RL) framework, which the agent uses with KG properties to predict the operations it wants to perform. In this paper, they used AI2-THOR framework to evaluate their approach. The reasoning used in this method is like human action in an environment to navigate action. Just as a human uses the semantic structure he already has in mind to find and identify an object in an environment, so does this approach. For example, we go to the refrigerator in the kitchen to find fruit. Agent knowledge is encrypted by the GCN algorithm in a graph. Agent knowledge is also updated according to the current observations it makes and the knowledge it has gained from previous steps.The steps of this approach are as follows:

**1.** Integrate a deep RL model with a KG

**2.** Use semantic prior knowledge for navigation

**3.** Using the previous information and the semantic structure that exists between objects. This approach can be generalized and used in unknown environments and new objects that are unknown[17]. Self-driving vehicles include different parts that can be categorized as follows[19]:

Recently, with the development of self-driving vehicles, the use of KG has become especially important. Since the amount of information received by sensors and cameras in a scene of driving is very large and this information is constantly updated, to use artificial intelligence and machine learning methods by which machine learning models can be taught. We need the received data to be properly structured so that we can better manage and organize them[14].Using semantic knowledge to organize and give meaning to the objects in the scenes, which constitute our data, can improve performance and speed up learning, and then perform and select the appropriate action to deal with factors in driving, such as pedestrian, dealing with obstacles, traffic signs, choosing a better and shorter route, as well as reducing traffic, etc. are very effective.Using semantic knowledge and then structuring the data obtained by the KG can help our agent (car) to perform better in an Reinforcement learning environment and perform better actions according to the knowledge it acquires[17][18].

According to [17], they focus on meanings using semantic priors in their work to navigate. They also used Graph Convolutional Networks to integrate prior knowledge into a deep Reinforcement learning (RL) framework, which the agent uses with KG properties to predict the operations it wants to perform. In this paper, they used AI2-THOR framework to evaluate their approach. The reasoning used in this method is like human action in an environment to navigate action. Just as a human uses the semantic structure he already has in mind to find and identify an object in an environment, so does this approach. For example, we go to the refrigerator in the kitchen to find fruit. Agent knowledge is encrypted by the GCN algorithm in a graph. Agent knowledge is also updated according to the current observations it makes and the knowledge it has gained from previous steps.

The steps of this approach are as follows:

**1.** Integrate a deep RL model with a KG

**2.** Use semantic prior knowledge for navigation

**3.** Using the previous information and the semantic structure that exists between objects. This approach can be generalized and used in unknown environments and new objects that are unknown [17]. Self-driving vehicles include different parts that can be categorized as follows 2.2 [19]:

| The structure of Autonomous vehicle | | Description |
|---|---|---|
| **Hardware** | Sensors | Lidar, millimeter radar, sonar, GPS, steering angle, velocity, accelerometer, camera (stereo, IR, FIR) |
| | Body | motor, steering, brake, accelerator, body material, shape, etc |
| **Software** | Object/position detection | lane, car, human, bicycle, etc |
| | Scene understanding | object movement, intension, collision detection |
| | Control | steering, speed, path planning |

Figure 2.2: Self-driving vehicles include different parts that can be categorized as Hardware and Software

Graph search-based methods are used to find shorter and more efficient routes in AVs. In these methods, the environment in which the car is located is considered as a graph, which includes many nodes through which the paths are connected together, then according to the current state of the car and the target state, Algorithms such as Dijkstra, A-star and other types of A-star are used to find better and shorter paths [20].The Dijkstra algorithm is a greedy algorithm that we can use to find the least distance or the least cost in a graph [21].

According to [20], The below articles have used the Dijkstra algorithm for routing in recent years.

**1.** Safe and Reliable Path Planning for the Autonomous Vehicle Verdino[22]

**2.** Odin: Team VictorTango's entry in the DARPA Urban Challenge[23]

**3.** Multi-Level Planning for Semi-autonomous vehicles in Traffic Scenarios Based on Separation Maximization [24]

The A-star algorithm is used to estimate the shortest path in real-world problems such as maps and environments where there may be many obstacles. A-star Algorithm defines a function f(n) for each node, which is an estimate of the total cost of a path

[1][25].

According to [20], The below articles have used the A-star algorithm for AVs:

**1.** Virginia Techs Twin Contenders: A Comparative Study of Reactive and Deliberative Navigation [26].

**2.** Navigating car-like robots in unstructured environments using an obstacle sensitive cost function [27].

**Autonomous vehicles perception**

Perception in AVs cars is very important to increase safety performance and reliability because vehicle decisions in different situations are based on the received data. Errors in the received data can cause irreparable human and financial losses. Various methods have been proposed for the perception of vehicles from their environments, which include important methods: localization, road mapping, detection of fixed and moving obstacles, traffic control, etc.[20]

According to [20], Localization methods that are not dependent on GPS can be classified into three categories:

**1.** LIDAR-based

**2.** LIDAR plus and camera-based

**3.** Camera-based

**Motion Planning Architecture**

According to [28], Motion planning includes two methods, Decoupled planning and direct planning.

Direct Planning method first tries to find an optimal path. Using the A-star algorithm, a path is found in a three-dimensional graph. In this method, the spatial distances of the nodes in the graph are adjusted and adapted by the speed of the vehicle while the time differences are constant. This is an online algorithm. This algorithm has a very limited search space and is therefore non-optimal.[29][28]

In Decoupled Planning method, motion planning is divided into different parts and then they are examined and solved sequentially. Because the problem is divided into subproblems, it will be easier to solve. As a result, Decoupled planning is less complex[30][28].

According to [31], the model presented in this paper is an abstract model that allows vehicles to understand and interact relative to roads and road traffic. This connection protocol between vehicles helps to resolve various road traffic conditions through negotiation. In this paper, by presenting a graph road that includes completely independent vehicles, they have been able to model the road and use it. This model has different elements of traffic, such as traffic flow, vehicle information and cars location. This paper consists of two parts, in the first part they are modelled and described by graphs, roads and vehicles and traffic, and in the second part they provide traffic control protocols. Traffic control protocols can be divided into two main categories:

1. Time-based traffic control

2. Priority-based traffic control

---

[1]https://brilliant.org/wiki/a-star-search/

The main application of time-based traffic control protocols is traffic light systems. Priority-based protocols control traffic based on pre-determined priorities, for example, the priority of vehicles at intersections or the priority of vehicles when bypassing the square, etc. All this information can be modelled and formulated by the graph structure and used for better interaction of AVs [31].

## 2.2 Segmentation for Autonomous Driving

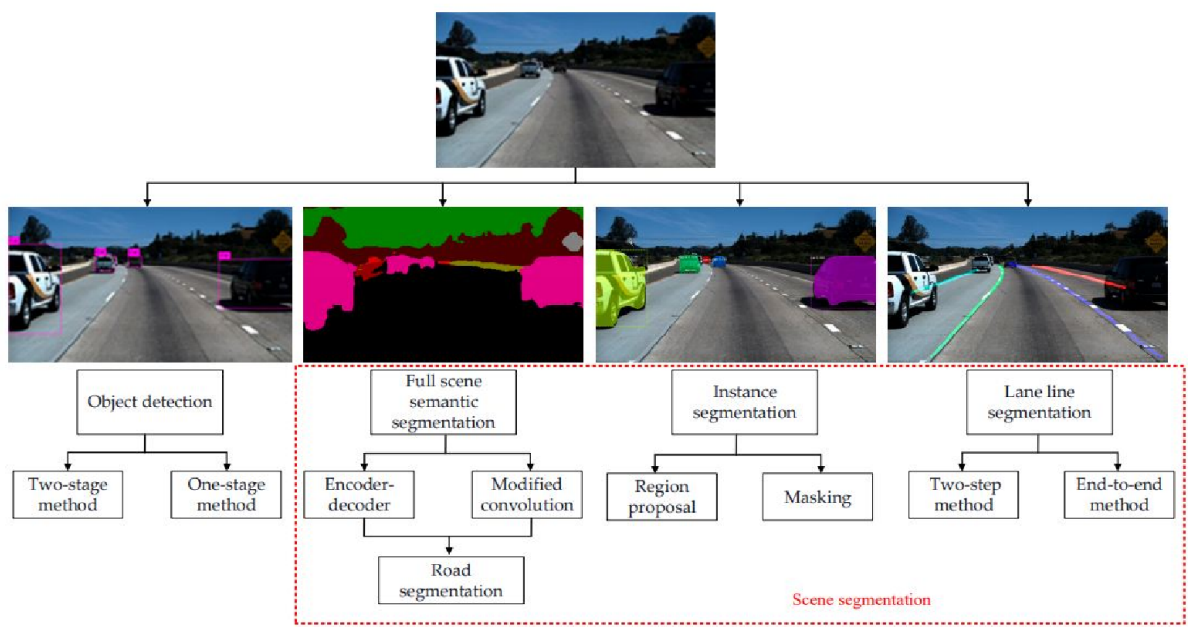

Figure 2.3: The research on scene understanding is organized as four work streams. [2]

In computer images, the term "image segmentation" or "segmentation" refers to the division of an image into a group of pixels based on certain criteria. An image segmentation algorithm captures the input and output of a set of regions (or segments). Effective segmentation of objects in a color image is an important issue for image processing operators. With effective segmentation, we separate the desired element. The superiority of one segmentation method over other methods depends on the specific characteristics of the problem under consideration. Color Image segmentation In many image processing tasks, such as image therapy, machine vision, image compression, objectology, is an essential need to start processing the object or texture. In therapeutic images, for example, a physician uses their own knowledge and experience to localize the tissue in the image. But when the number of images is large, or when the contrast and change in brightness of objects relative to each other is low and the image is unsuitable from a human point of view, segmentation is very costly (both financially and temporally) and with error. Therefore, the need to automate the image segmentation process is necessary. Image segmentation is done in different ways that can be generally divided into two categories: classical and morphological. Semantic segmentation of images means estimating the class for each pixel of the image. Image segmentation is the process of dividing an image into several parts. In this process, each pixel of the image is assigned to an object. The two main methods of image segmentation are semantic segmentation and instance segmentation. In semantic segmentation, all objects that are of the same type are marked with a label and placed in a class; But in instance image segmentation, similar objects are separated from each other and each will receive a separate tag.

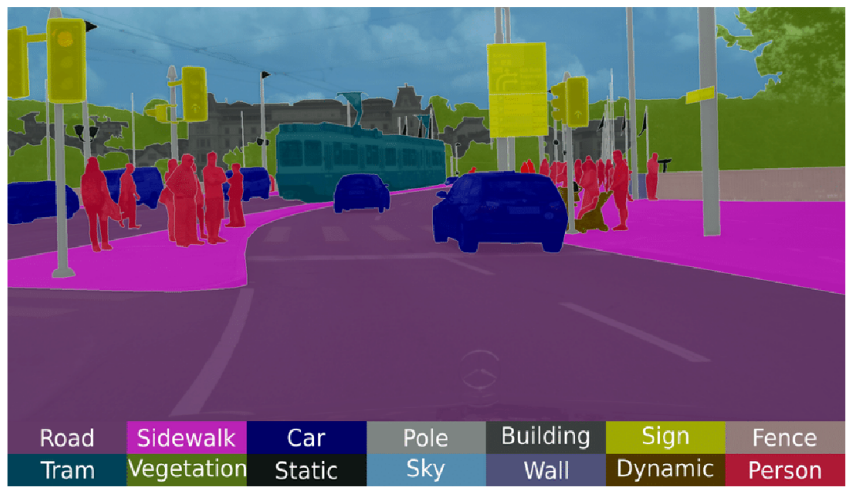

Figure 2.4: Semantic segmentation of a scene from the Cityscapes dataset [3][4].

The main architecture and structure in the image segmentation include an encoder and a decoder. The encoding section extracts the specific properties of each image using filters. The decoder is also responsible for generating the final output, in which a segmentation mask typically outlines the object. The architecture of image segmentation processes is, in most cases, similar to the architecture of the figure below. Full-scene se-mantic segmentation is the division of groups of objects at the pixel level into a complete image. Full-scene semantic segmentation approaches fall into two categories: encoder-decoder structure models and modified convolution structure models. As a special and important task, scene understanding in autonomous vehicles segmentation has been extensively studied.

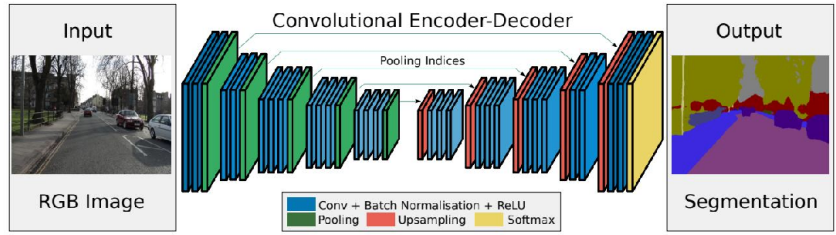

Figure 2.5: An illustration of the SegNet architecture [5].

Figure 2.6 gives a summary of the representative works in terms of their characteristics, core technology and functions, basic frameworks [2].

| Method Category | Typical Work | Characteristics | Core Technology and Functions | Basic Framework |
|---|---|---|---|---|
| Encoder–Decoder | FCN SegNet U-Net ENet PSPNet Deeplab-V3+ Fast FCN CED-Net DANet | End-to-end dense pixel output The pyramid pooling module can ensure global information integrity | (1) Deconvolution: Upsampling (2) UnPooling: Increasing the resolution of the feature map (3) Bilinear interpolation: Restoring the image Size. | |
| Modified Convolution | Deeplab-V1 Dilated convolution Deeplab-V2 Deeplab-V3 CRFasRNN DRN HDC Deeplab-V3+ | Ensure local information correlates through modified convolution | (1) Dilated convolution: Increasing convolution receptive fields (2) ASPP: Capturing image global information | |

Figure 2.6: Comparison of deep learning-based approaches for full-scene semantic segmentation.

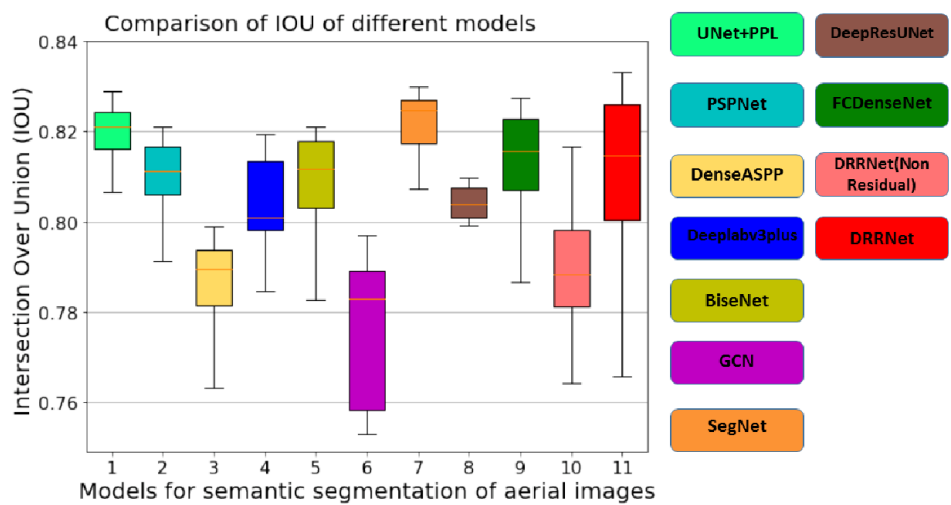

Figure 2.7: Box plot of Intersection Over Union of different models [6].

Figure 2.7 compares the box plot of Intersection Over Union of different models.

## 2.3  Review of loss functions for semantic segmentation

In this section, we summarize some of the well-known loss functions that are widely used for image segmentation and list those that can help a model converge faster and better. Image segmentation can be defined as a pixel-level classification task. An image is made up of different pixels, and these pixels together define different elements in the image. A method for classifying these pixels into elements is called semantic image segmentation. The choice of loss/goal function is very important when designing deep learning architectures based on image segmentation because they stimulate the learning process of the algorithm.Since 2012, researchers have experimented with a variety of domain-specific loss functions to improve results for their datasets. In this section, we will review some of the segmentation-based loss functions that are important. These loss functions can be divided into 4 categories: distribution-based, area-based, boundary-based, and Compounded-based. We also discuss the situation to determine which goal/loss function might be useful in a scenario.

**Binary Cross-Entropy**

Cross-entropy is a mathematical method used in discrete problems such as classification. This function calculates the distance between two probabilities. It's extensively used for classification purposes, and it works well because segmentation involves pixel-level categorization [32].

Binary Cross-Entropy is defined as:

(2.1)
$$L_{BCE}(y, \hat{y}) = -(y \log(\hat{y}) + (1 - y) \log(1 - \hat{y}))$$

here, $y$ is true value, and $\hat{y}$ is the predicted outcome.

If $M > 2$ (i.e. multiclass classification), we calculate a separate loss for each class label per observation and sum the result.

(2.2)
$$-\sum_{c=1}^{M} y_{o,c} \log(\hat{y}_{o,c})$$

$M$ - number of classes

$log$ - the natural log

$y$ - binary indicator (0 or 1) if class label c is the correct classification for observation o

$\hat{y}$ - predicted probability observation o is of class c

| Type | Loss Function |
|------|---------------|
| Distribution-based Loss | Binary Cross-Entropy |
| | Weighted Cross-Entropy |
| | Balanced Cross-Entropy |
| | Focal Loss |
| | Distance map derived loss penalty term |
| Region-based Loss | Dice Loss |
| | Sensitivity-Specificity Loss |
| | Tversky Loss |
| | Focal Tversky Loss |
| | Log-Cosh Dice Loss |
| Boundary-based Loss | Hausdorff Distance loss |
| | Shape aware loss |
| Compounded Loss | Combo Loss |
| | Exponential Logarithmic Loss |

Table 2.1: TYPES OF SEMANTIC SEGMENTATION LOSS FUNCTIONS

**Dice Loss**

A widely used metric in computer vision, the Dice coefficient is used to determine the similarity of two images. In 2016, it was also adapted into a loss function called Dice Loss.

$$(2.3) \qquad\qquad DL(y, \hat{p}) = 1 - \frac{2y\hat{p} + 1}{y + \hat{p} + 1}$$

The function here is defined by including 1 in the numerator and denominator to ensure that it is not undefined in edge case scenarios, such as when $y = \hat{p} = 0$ [32].

**Mean Squared Error(MSE)**

Is one of the most well-known loss functions in regression, and calculates the average squared difference between the actual and predicted values by the following equation:

$$(2.4) \qquad\qquad MSE(y, \hat{y}) = \frac{1}{n} \sum_{i=1}^{n} (y_i - \hat{y}_i)^2$$

## 2.4 Review of activation functions in semantic segmentation

The activation function plays a very important and key role in the architecture of a neural network model. This function is used to propagate the output of each layer to the other at the end of the computational process in each neuron. In simple terms, the activator function is responsible for deciding which neurons should be activated or which should be inactivated. In general, nonlinear activation functions are more commonly used in neural networks. The activation function used in feedforward networks, unlike some other networks, can not be of any function but must have certain features. This function must be continuous, derivative, and uniformly descending, also the first derivative of this function must be easily computable. In the following, we will review some examples of widely used activating functions in semantic segmentation.

**Sigmoid**

The Sigmoid function, also known as the logic function, is one of the most useful nonlinear activation functions in artificial neural networks. This function is used to calculate the probability for binary classification problems in the output layer. This function generates probabilistic output in the form of values between zero and one for each category. We use this function in Binary semantic segmentation.

The sigmoid function is defined as:

$$\sigma(x) = \frac{1}{1 + e^{-x}} \tag{2.5}$$

**Rectified Linear Unit (ReLU)**

The ReLU function, used in the hidden layer, is one of the most widely used functions in deep learning today. By setting the negative input values to zero, it converts the input to a value greater than or equal to zero. In other words, it has no upper limit for positive inputs [33].

The ReLU function is defined as:

$$Relu(x) = max(0, x) \tag{2.6}$$

**Softmax**

This function, which is used in the output layer, is a generalization of the Sigmoid activation function and is used for classification problems. For classification problems with more than two categories, this function makes it possible to make a probabilistic prediction. We use this function in Multiclass semantic segmentation.

The Softmax function is defined as:

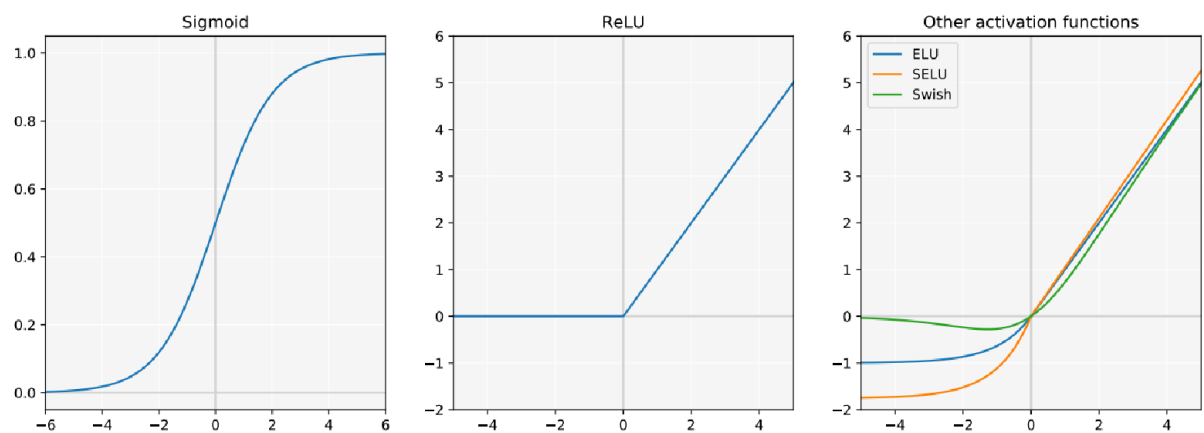

Figure 2.8: Many different activation functions act as non-linearities in the network. Sigmoid and ReLU are the most commonly used, although ELU, SELU, and Swish are also developed [7].

(2.7)
$$\sigma(x_i) = \frac{e^{x_i}}{\sum_{j=1}^{K} e^{x_j}} \quad for \ i = 1, 2, \ldots, K$$

| Function | Domain | Zero-axis | Saturation Problem | Vanishing Gradients Problem | Computing |
|---|---|---|---|---|---|
| Sigmoid | [0,1] | No | For positive and negative values | Yes | Slow and exponential |
| Tanh | [-1,1] | Yes | For positive and negative values | Yes | Slow and exponential |
| ReLU | [0,+∞] | No | For negative values | Better than Sigmoid and Hyperbolic Tangent | Fast |
| LeakyReLU | [−∞,+∞] | Yes | No | No | Fast |

Table 2.2: Comparison of activation functions

## 2.5 Layers of the Neural Network

Often referred to as layers, deep neural networks consist of multiple stacked, differentiable modules. This section describes the most popular layers in neural networks.

### 2.5.1 Fully Connected Layers

The neurons in a fully connected layer are all connected to each other, just like in a regular Neural Network, so their activations can be computed using a matrix multiplication followed by a bias offset. Basically, this is a layer with learnable parameters that linearly transforms the output of the previous layer, and then squashes the result with a non-linear activation function, such as a sigmoid or a ReLU: $y = f(Wx + b)$.

A fully connected layer has a large weight matrix $W$: it connects each element of $x$ (activations of the previous layer) with each element of $y$ (activations of the current layer). Figure 2.9 [2] illustrates this pattern of connectivity. This large weight matrix introduces a large number of parameters into the model and could lead to overfitting.

### 2.5.2 Locally Connected Layers

Rather than connecting x and y densely, a locally connected layer connects them sparsely: each neuron in the current layer is only connected to neurons in the previous layer that are in the same spatial neighborhood. Data with a spatial structure, such as images, benefit from the connectivity pattern with a local receptive field because pixels are typically locally correlated in the spatial dimensions. A locally connected layer introduces fewer model parameters than a fully connected layer for the same dimensions of x and y.

---

[2]https://developer.nvidia.com/deep-learning

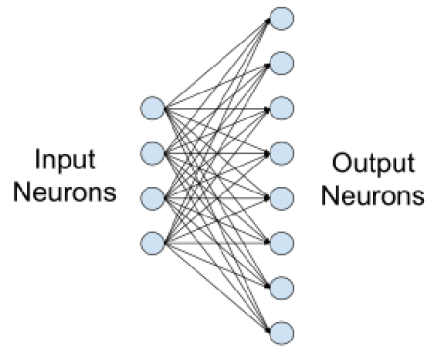

Figure 2.9: Example of a small fully-connected layer with four input and eight output neurons.

### 2.5.3 Convolutional Layers

The Convolutional layer is the most important part of CNN and is always used as the first layer. This layer is responsible for most of the computational load. In general, CNN is a neural network that has at least one convolutional layer in its structure. Convolutional, in its most general definition, is the performance of mathematical operations on two functions with real values. The main task of the convolutional layer is to identify the features found in the local areas of the input image, which are common to the entire data set. This feature identification leads to the production of a feature map by applying filters. The convolutional layer applies a local filter to the input image. This results in better classification of neighboring pixels that are more correlated in the same image. In other words, the pixels of the input images can be correlated with each other. For example, in facial images, the nose is always between the eyes and the mouth. When we apply the filter to subsets of the image, we extract some local properties.

### 2.5.4 Pooling Layers

The Pooling layer is usually used periodically between two consecutive layers of convolutional. Its task is to reduce the size of feature maps. In addition to extracting important features in the feature map, this also reduces the computational power required to process the data by reducing the number of parameters. There are two important pooling layers: max pooling and average pooling. max-pooling has better performance in extracting dominant and important features.

### 2.5.5 Batch Normalization

One of the problems in neural network training, in addition to vanishing gradient and gradient explosion, is the problem of changing the internal variables of the network. This problem arises because the parameters are constantly changing during the training process, which in turn changes the values of the activation functions. Changing the input values from the initial layers to the subsequent layers causes a slower convergence during the training process because the training data of the subsequent layers are not stable. Batch Normalization is proposed to overcome this problem to reduce instability and improve the network. In this method, it performs batch normalization on the input data of a layer in such a way that they have a mean of zero and a standard deviation of one. This simplifies the learning process in the model, as the parameters in the previous layers will in most cases be ineffective. Without this batch normalization, each update will make a drastic change to the model. In summary, by placing a batch normalization between the hidden layers and creating a common variance property, we reduce the internal changes of the network layers.

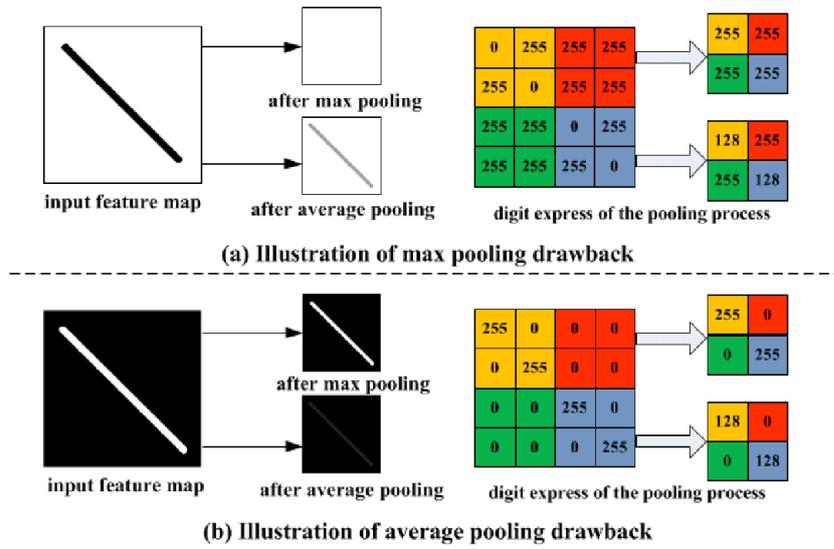

Figure 2.10: Example of max pooling VS average pooling [8].

REVIEW OF DATASETS FOR AUTONOMOUS VEHICLES

## 3.1 Datasets

### 3.1.0.1 nuScenes

The nuScenes dataset is a public large-scale dataset for autonomous driving developed by the team at Motional (formerly nuTonomy). The dataset includes 1,000 driving scenes in two busiest cities of Boston and Singapore. nuScenes is the first large-scale data set that collects data from the entire set of sensors of an independent vehicle including 6 Cameras, 1 LIDAR, 5 RADAR, GPS, IMU [1][34]. The authors present nuScenes dataset, detection and tracking tasks, metrics, baselines, and results in this paper [34]. AVs are tested on public roads for the first time, and this is the first dataset that includes the full 360°sensor suite (lidar, images, and radar). Among all previously released datasets, nuScenes has the most annotations of 3D boxes. In order to stimulate research on 3D object detection for autonomous vehicles, a new detection metric balancing all aspects of detection performance is introduced. A number of leading lidar and image object detectors and trackers are demonstrated on nuScenes in novel adaptations. In addition, the researchers plan to add image-level and point-level semantic labels and a benchmark for trajectory prediction in the future [34].

---

[1]https://www.nuscenes.org/

### 3.1.0.2 Lyft

The dataset includes over 55,000 human-labelled 3D annotated frames, surface map, and an underlying HD spatial semantic map that is captured by 7 cameras and up to 3 Lidar sensors.This database uses nuScenes [2]. For training prediction and planning solutions, this paper [35] presents the largest and most detailed dataset available. As compared with the current best alternatives, this dataset is three times larger and significantly more descriptive. Their results suggest that both motion forecasting and motion planning performance increase as a result of this difference. Large-scale machine learning systems are highly dependent on datasets, which is consistent with intuition. As they are often the result of proprietary industrial efforts, these datasets are not available to everyone. The publication of this dataset represents an important step toward democratizing self-driving applications. Through the use of this dataset, a fully autonomous future can be reached more rapidly. In addition, they observe that motion forecasting and motion planning performance increase with training data size. Therefore, even larger datasets with tens of thousands or even millions of hours, together with algorithms that utilize them, will be desirable in the future [35].

### 3.1.0.3 Visual Genome

Visual Genome is a dataset, a knowledge base, an ongoing effort to connect structured image concepts to language [3]. A multi-layered understanding of pictures is provided by Visual Genome. From pixel-level details, such as objects, to relationships that require further explanation, and even to cognitive tasks such as answering questions, it allows for multi-perspective analysis of images. Computer vision models can be trained and benchmarked using this dataset. As a result of the Visual Genome, computers will be able to detect objects and describe them, as well as explain their interactions and relationships, enabling them to develop a better understanding of the visual world. A Visual Genome is a formalized knowledge representation that grounds visual concepts in language and provides a detailed set of descriptions and question answers [36].

This dataset includes the following:

**1.** 108,077 Images

**2.** 5.4 Million Region Descriptions

**3.** 1.7 Million Visual Question Answers

**4.** 3.8 Million Object Instances

---

[2]https://self-driving.lyft.com/
[3]https://visualgenome.org/

**5.** 2.8 Million Attributes

**6.** 2.3 Million Relationships

**7.** Everything Mapped to Wordnet Synsets [36]

#### 3.1.0.4 Common Objects In Context Dataset

COCO is a large-scale object detection, segmentation, and captioning dataset. The authors of this paper [37] developed a new dataset for detecting and segmenting everyday objects in their natural environment. An extensive collection of object instances was collected, annotated, and organized to drive the advancement of object detection and segmentation algorithms using over 70,000 worker hours. The objective was to find images of objects in natural environments, from varied perspectives, that were not iconic. According to the dataset statistics, each image contains a wealth of contextual information with a variety of objects [37].

COCO has several features:[4]

**1.** Object segmentation

**2.** Recognition in context

**3.** Superpixel stuff segmentation

**4.** 330K images (more than 200K labeled)

**5.** 1.5 million object instances

**6.** 80 object categories

**7.** 91 stuff categories

**8.** 5 captions per image

**9.** 250,000 people with keypoints

#### 3.1.0.5 The Cambridge-driving Labeled Video Database (CamVid)

The CamVid Database provides ground truth labels that associate each pixel with one of 32 semantic classes. In order to achieve the long-term goals of object analysis research, objects must be able to be identified even in motion. They propose the CamVid annotated database as a means of evaluating and improving these object recognition algorithms. They developed this database as a direct response to the formidable challenge of segmenting video data semantically. Research on object analysis can benefit from four contributions made by the CamVid Database. As a first step, the per-pixel class labels for at least 700 images at 1 or 15Hz provide the first ground truth for video-based multiclass object recognition. Two humans agreeing on each frame improves the reliability of the labels. Further, the large-resolution and high-quality video images filmed at 30 frames per second provide valuable extended duration footage regarding driving scenarios and ego-motion. Additionally, they were able to show camera calibration data and 3D pose tracking data based on the controlled conditions under which each frame was filmed. It would be ideal if algorithms did not need this information, but higher-level object analysis should not be impeded by the sub-task of auto-calibration. As a final feature, the database comes with custom-made software for users who wish to label other images and videos precisely [38].

---

[4]https://cocodataset.org/

The dataset is used in semantic segmentation research.

This database includes the following:

**1.** 367 training pairs

**2.** 101 validation pairs

**3.** 233 test pairs

[5][38][39][18]

| | Number of Labelled Images | Classes | Multiple Cities | Environment |
|---|---|---|---|---|
| **KITTI** | **200** | **34** | **No** | **Daylight** |
| **Cityscapes** | **3478** | **34** | **Yes** | **Daylight** |
| **Mapillary** | **20k** | **66** | **Yes** | **Daylight, rain, snow, fog, haze, dawn, dusk and night** |
| **Apollo Scape** | **147k** | **36** | **No** | **Daylight, snow, rain, foggy** |
| **BDD100K** | **8000** | **19** | **Yes** | **Daylight, rain, snow, fog, haze, dawn, dusk and night** |

Figure 3.1: Publicly available datasets for urban driving scenes (Autonomous Driving)

---

[5]http://mi.eng.cam.ac.uk/research/projects/VideoRec/CamVid/#ClassLabels

## SCENE UNDERSTANDING METHODS AND IMPLEMENTATION DETAILS

# 4.1 Segmentation Models

In order to understand the environment in our proposed method, a segmentation architecture is required. This section discusses the models proposed for this project and then presents the model architecture that had the best results.

## 4.1.1 Unet

The U-Net [40] is one of the most famous Fully Convolutional Networks (FCN) in biomedical image segmentation, which was published in 2015 by MICCAI with more than 40000 citations. Using the network they modify a bit to segment the dental X-ray image in IEEE International Symposium on Biomedical Imaging (ISBI) 2015. They also segment the electron microscopic image in this paper.

Figure 4.1 depicts the network architecture, which is composed of a contracting path (left) and an expansive path (right). Contracting paths follow the typical architecture of a convolutional network.The process consists of applying two 3x3 convolutions (unpadded convolutions), each followed by a rectified linear unit (ReLU) and a pooling operation using $2 \times 2$ max with stride 2 for downsampling. The number of feature channels is doubled at each downsampling step. As each step of the expansive path occurs, the feature map is upsampled followed by a $2 \times 2$ convolution ("up-convolution") that halves

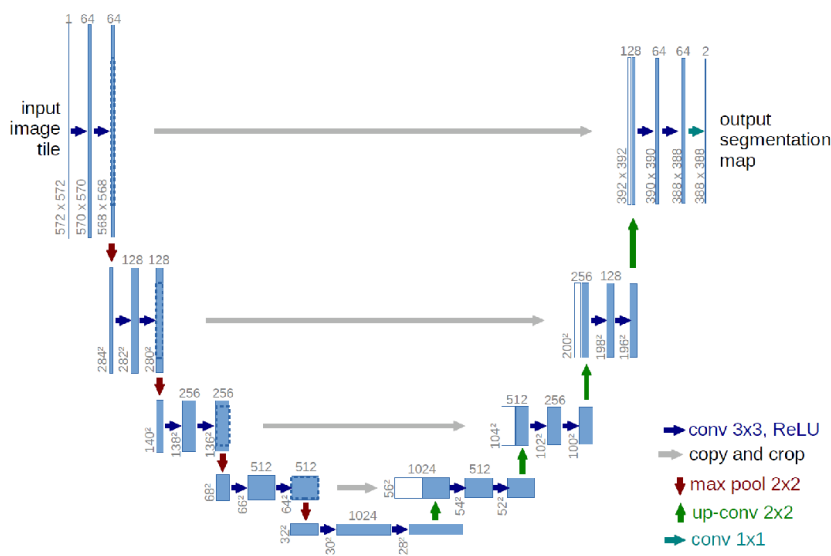

Figure 4.1: U-Net architecture (for 32×32 pixels at the lowest resolution). Each blue box
represents a multi-channel feature map. The number of channels appears on the box.
A description of the x-y size appears on the lower-left side of each box. There are white
boxes representing copied feature maps, and arrows denote the different operations.

the number of channels of the feature map. Two 3 × 3 convolutions followed by a ReLU
are added to the concatenation with the cropped feature map from the contracting path.
The cropping is required because of the loss of border pixels in every convolution. At
the final layer, 64 component feature vectors are mapped to an appropriate number of
classes using 1 × 1 convolution. The network has a total of 23 convolutional layers [40].

## 4.1.2   FPN

Using ConvNet's pyramidal feature hierarchy, which has semantics from low to high,
they intend to build an entire feature pyramid using semantics at a high level. Essen-
tially, the result is a general-purpose Feature Pyramid Network. The FPNs are also
generalized to instance segmentation proposals. As explained in the following, our pyra-
mid is built by using bottom-up pathways, top-down pathways, and lateral connections
[41].

**Bottom-Up Pathway**

Feedforward computation of the backbone ConvNet is the bottom-up pathway. For
each stage, one pyramid level is defined. The output of each stage's last layer serves as
a reference set of feature maps that we will enrich to create our pyramid. As a result,
the deepest layer of each stage has the strongest features.

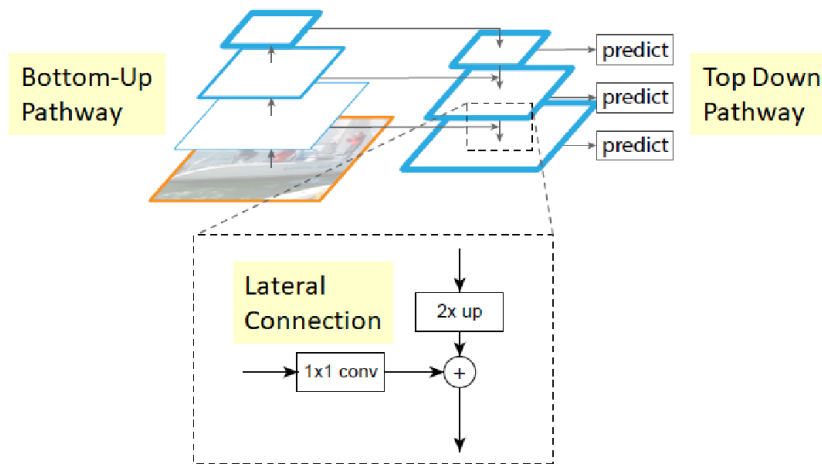

Figure 4.2: Feature Pyramid Network (FPN)

**Top-Down Pathway and Lateral Connection**

- By upsampling spatially coarser but semantically stronger feature maps from higher pyramid levels, the top-down pathway presents higher resolution features. For simplicity, the spatial resolution is upsampled by a factor of two while using the nearest neighbor.

- The lateral connections merge same-sized feature maps from bottom-up and top-down pathways.

- In particular, to reduce channel dimensions, the feature maps from the bottom-up pathway undergo $1 \times 1$ convolutions.

- Feature maps from the top-down pathway and the bottom-up pathway are merged by adding element-wise.

**Segmentation**

As with Mask R-CNN, FPN is also effective at extracting masks for image segmentation. With MLP, a window of dimensions $5 \times 5$ is slide over each feature map to generate object segments of dimensions $14 \times 14$. After merging masks at a different scale, we form our final mask predictions [41].

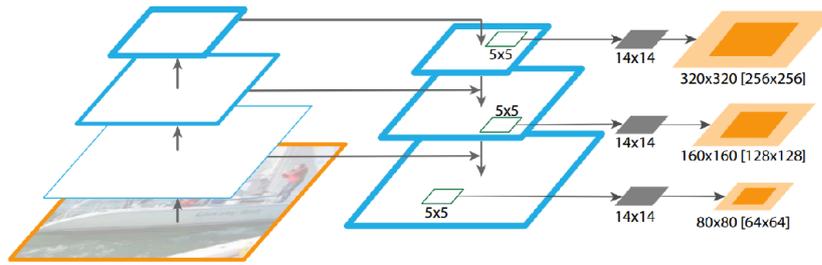

Figure 4.3: Feature Pyramid Network (FPN) for object segment proposals.

### 4.1.3 Linknet

For visual scene understanding to be useful in real-time applications, pixel-wise seman-
tic segmentation needs to be accurate as well as efficient. Even though existing algo-
rithms are accurate, they do not focus on utilizing neural network parameters to their fullest potential. The result is that they are huge in terms of parameters and number of operations, so they are also slow. In paper [42], the authors suggest a novel architecture for deep neural networks that permits them to learn without a significant increase in parameter count.

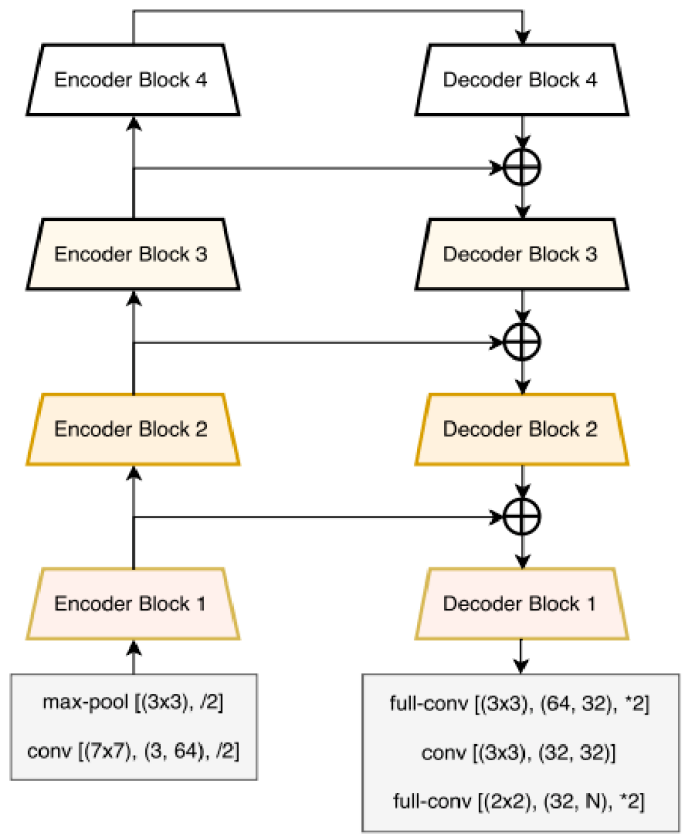

Figure 4.4: LinkNet Architecture

Figure 4.4 illustrates the architecture of LinkNet. Conv and full-conv refer to convolution and full-convolution, respectively. Additionally, /2 denotes downsampling by a factor of 2, which is carried out using strided convolution, and *2 means upsampling by a factor of 2. Between each convolutional layer, batch normalization is applied, followed by ReLU non-linearity. In Figure 4.4, the left half of the network is the encoder, while the right half is the decoder. In the encoder, starting from an initial block, convolution is conducted on the input image with a kernel of size 7 × 7 and a stride of 2. A spatial max-pooling procedure is also performed here, with an area of 3×3 and a stride of 2. The later part of encoder consists of residual blocks which is represented as encoder-block(i). Figure 4.5(a) illustrates in detail the layers within these encoder-blocks. In Figure 4.5(b), layer details for decoder-blocks are also provided [42].

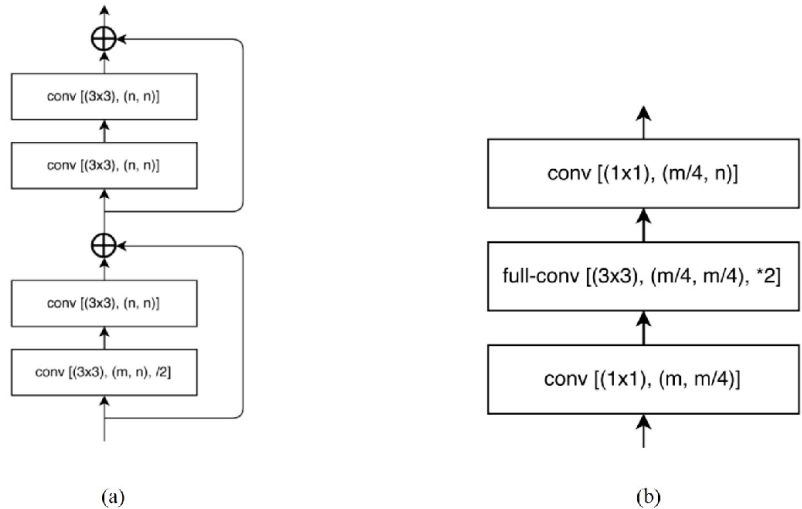

Figure 4.5: (a) Convolutional modules in encoder-block (i), and (b) Convolutional modules in decoder-block (i).

### 4.1.4   PSPNet

**Pyramid Pooling Module**

Figure 4.6 illustrates a description of PSPNet. At (a), we have an input image. To extract features at (b), the ResNet technique is combined with a dilated network strat-egy (DeepLab / DilatedNet). The dilated convolution is following DeepLab. We have a
feature map that is 1/8 the size of the input image.

### (c).1. Sub-Region Average Pooling

A sub-region average pooling for each feature map is applied in (c).

- **Red:** The coarsest level, highlighted in red, is a single bin output generated from
  global pooling.

- **Orange:** In this second level, the feature map is divided into 2 × 2 sub-regions,
  and average pooling is done for each sub-region.

- **Blue:** The third level divides the feature map into 3 × 3 sub-regions and then
  performs average pooling for each.

- **Green:** At this level, the feature map is divided into 6×6 sub-regions, then pooling
  is applied to each sub-region.

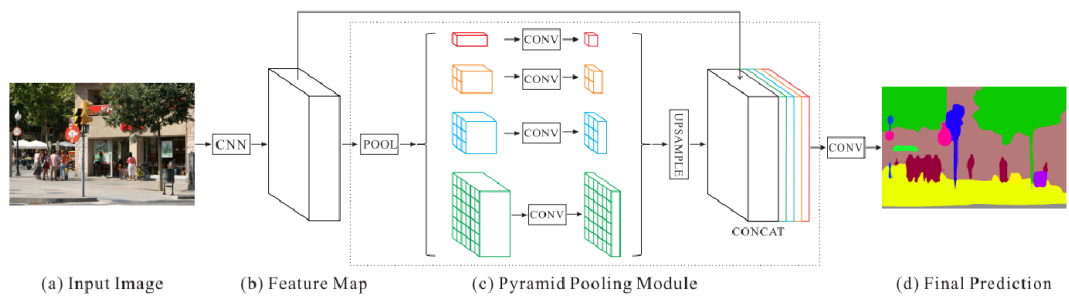

Figure 4.6: A description of PSPNet. Assuming an input image (a), First, we use CNN
to obtain the feature map of the last convolutional layer (b), In (c), the final feature
representation is formed by combining multiple subregion representations and applying
a pyramid parsing module to harvest different subregion representations. Upsampling
and concatenation layers are then applied to form the final feature representation which
includes both local and global context [9].

### (c).2. $1 \times 1$ Convolution for Dimension Reduction

If the level size of the pyramid is N, then $1 \times 1$ convolution is performed for each
pooled feature map to reduce the context representation to $1/N$ of the original one
(black), maintaining the weight of the global feature.

### (c).3. Bilinear Interpolation for Upsampling

Each low dimension feature map is up-sampled using bilinear interpolation in order
to have the same dimensions as the original feature map (black).

**(c).4. Concatenation for Context Aggregation**

The original feature map (black) is concatenated with all levels of upsampled feature maps. As a global prior, these feature maps are fused. At (c), the pyramid pooling module is completed.

**(d)**

In the end, a convolution layer is applied to generate the final prediction map at (d) [9].

## 4.1.5 DeepLabv3+

Figure 4.8 illustrates improved DeepLabv3 Architecture.

- **a:** Through Atrous Spatial Pyramid Pooling (ASPP), it is possible to encode multi-scale contextual information.

- **b:** Location or spatial information can be recovered with Encoder-Decoder Architecture.

- **c:** (a) and (b) are utilised in DeepLabv3+.

### 4.1.5.1 Atrous Convolution

We apply atrous convolution over the input feature map x and the output region y for each location i and filter w, whereas the atrous rate r corresponds to the stride with which we sample the input signal [10].

$$(4.1) \qquad\qquad y[i] = \sum_k x[i + r.k] w[k]$$

### 4.1.5.2 Atrous Depthwise Convolution

The depthwise convolution supports atrous convolution. Furthermore, it is found that the computation complexity of the proposed model is significantly reduced while retaining similar (or better) performance [10].

### 4.1.5.3 DeepLabv3 as Encoder

In the context of image classification, the spatial resolution of the final feature maps is usually 32 times smaller than the input image resolution, which means that output

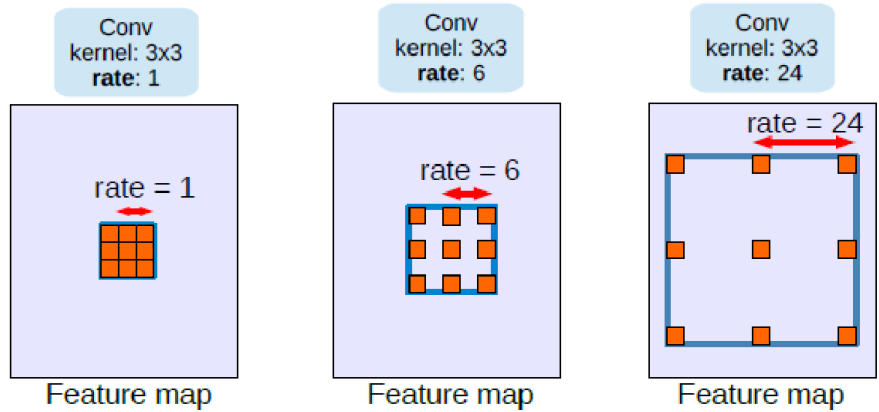

Figure 4.7: Atrous convolution with kernel size $3 \times 3$ and different rates. The standard convolution corresponds to an Atrous convolution of rate = 1. Large atrous rates enable object encoding at multiple scales by expanding the model's field of view [10].

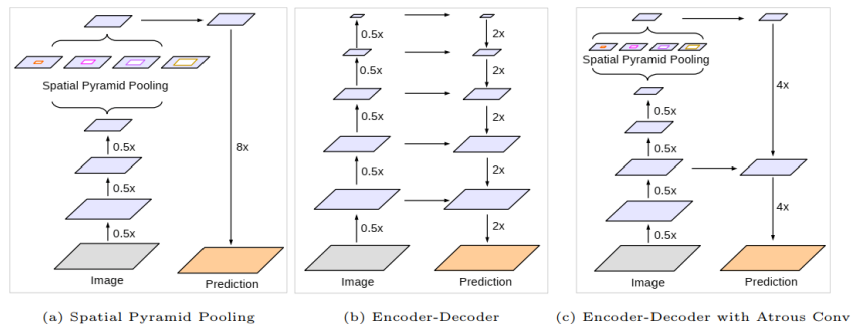

Figure 4.8: This figure illustrates improved DeepLabv3 Architecture [11].

stride = 32. It is too small for semantic segmentation. To obtain a denser feature extraction, output stride can be set to 16 (or 8) after removing the striding in the last block (or two) and applying the atrous convolution in that block(s).

### 4.1.5.4 Decoder

Bilinearly upsampling the encoder features by four times and concatenating them with low level features are the first steps. It is necessary to reduce the number of channels before concatenation by 1*1 convolution, since low-level features usually contain a large number of channels (e.g., 256 or 512), which may outweigh the importance of rich encoder features. To refine the features after concatenation, they apply a few $3 \times 3$ convolutions, followed by another simple bilinear upsampling by a factor of four [11].

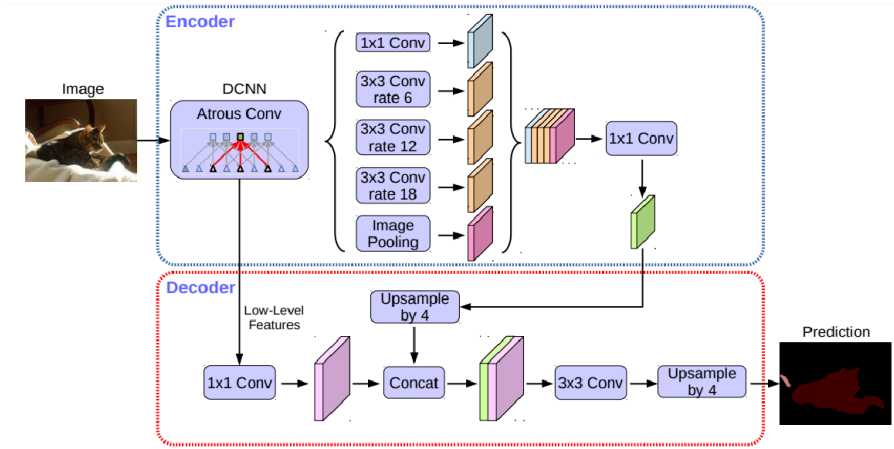

Figure 4.9: Using an encoder-decoder structure, DeepLabv3+ extends DeepLabv3 [11].

# 4.2 Implementation Details

In this section, we will discuss topics regarding implementation and design choices, such as libraries and frameworks, and topics pertaining to implementation of training.

## 4.2.1 Tensorflow and Keras

The networks and the training have been implemented using Tensorflow. There are many advantages to using Tensorflow for creating and running computational graphs. Multiple programming languages are supported by the API, and the core library is compiled in lower-level languages in order to generate optimized code for both CPUs and GPUs. We use the Python API in this thesis, which Tensorflow offers the most robust support for. In this section, we will study the fundamentals of Tensorflow and use Python's library of Neural Networks for Image Segmentation using Keras and Tensor-Flow.

## 4.2.2 Different Backbones for Semantic Segmentation Network

We used different backbones to perform this experiment. We tested each of them on 4 different models and then compared their mean Intersection Over Union (mIou) criteria to evaluate their performance. In other words, we examined which of the modes would have a better effect on scene understanding and semantic segmentation. We used different versions of backbones such as ResNet [43], VGG [44], DenseNet [45], Inception [46], MobileNet [47], and EfficientNet [48] to perform this experiment.

### 4.2.3 EfficientNet

The accuracy of models can be improved by scaling up their network dimensions, according to previous studies. Many of them, however, only tried to scale one dimension, such as depth, width, or resolution, while others tried to arbitrarily scale two or three dimensions, which require technical tuning and often yield suboptimal results. By rethinking the scaling process, the paper [48] tries to ascertain if there is a principled approach for scaling up Convents that would be more accurate and efficient.

There are two challenges in scaling up the dimensions. First and foremost, scaling up any dimension of network depth, width, or resolution improves accuracy, but its gain diminishes for larger models. The second important aspect of ConvNet scaling is to balance all dimensions of the network width, depth, and resolution. Compound scaling is a principled method for uniformly scaling network width, depth, and resolution [48].

### 4.2.4 Transfer learning

Transfer learning is a type of machine learning in which a model is first trained in a specific task, then some or all of the model is used as a starting point for a related task. To put it another way, we want to utilize the knowledge gained by a source task to assist in the learning of another target task. The purpose of transfer learning is to improve the process of learning new tasks by using the experience gained from solving previous problems that are somewhat similar. Figure 4.10 illustrates example of Transfer learning.

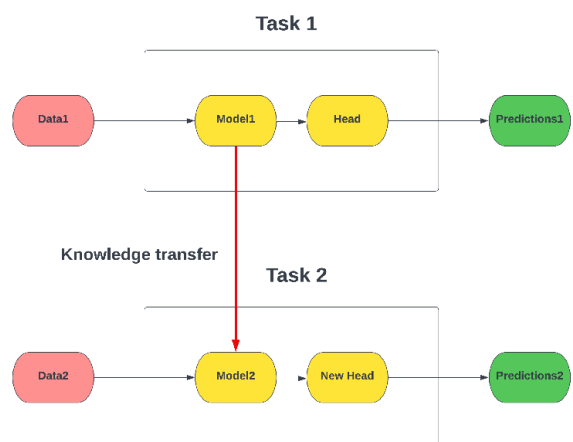

Figure 4.10: Example of Transfer learning.

# 4.3 Data preparation

In semantic segmentation, we need to first identify the classes in the masks to work with each dataset and consider a class for the pixels with the same colors. So we should replace RGB values with integer values to be used as labels. The figure below illustrates how to do this.

```
1   unlabeled = np.array([0,0,0]) #0
2   static_l = np.array([0,0,0]) #1
3   road = np.array([128, 64,128]) #2
4   sidewalk = np.array([244, 35,232]) #3
5   building = np.array([70, 70, 70]) #4
6   fence = np.array([190,153,153]) #5
7   wall = np.array([102,102,156]) #6
8   polegroup = np.array([153,153,153]) #7
9   traffic_light = np.array([250,170, 30]) #8
10  traffic_sign = np.array([220,220, 0]) #9
11  terrain = np.array([152,251,152]) #10
12  vegetation = np.array([107,142, 35]) #11
13  sky = np.array([70,130,180]) #12
14  person = np.array([220, 20, 60]) #13
15  rider = np.array([255, 0, 0]) #14
16  bicycle = np.array([119, 11, 32]) #15
17  bus = np.array([0,60,100]) #16
18  car = np.array([0,0,142]) #17
19  caravan = np.array([0, 0, 90]) #18
20  motorcycle = np.array([ 0, 0,230]) #19
21  train = np.array([0, 80,100]) #20
22  truck = np.array([0, 0, 70]) #21
```

Listing 4.1: Example of replace RGB values with integer values to be used as labels.

If the colors of the masks are hexadecimal, they can be converted to RGB as follows.

```
1   '''
2   RGB to HEX: (Hexadecimel --> base 16)
3   0-9 --> 0-9
4   10-15 --> A-F
5   Example: RGB --> R=201, G=, B=
6   R = 201/16 = 12 with remainder of 9. So hex code for R is C9 (
      remember C=12)
7   Calculating RGB from HEX: #3C1098
8   3C = 3*16 + 12 = 60
9   10 = 1*16 + 0 = 16
```

```
10    98 = 9*16 + 8 = 152
11    '''
12    #in python
13    a=int('3C', 16)
14    print(a) #3C with base 16. Should return 60.
```

Listing 4.2: Example of Convert HEX to RGB array.

We use the following function to find the pixel with the RGB combination for the arrays defined above. If it matches, replaces all values in that pixel with a specific integer.

```
1    def rgb_to_2D_label(label):
2
3        label_seg = np.zeros(label.shape,dtype=np.uint8)
4        label_seg [np.all(label == unlabeled,axis=-1)] = 0
5        label_seg [np.all(label==static_l,axis=-1)] = 1
6        label_seg [np.all(label==road,axis=-1)] = 2
7        label_seg [np.all(label==sidewalk,axis=-1)] = 3
8        label_seg [np.all(label==building,axis=-1)] = 4
9        label_seg [np.all(label==fence,axis=-1)] = 5
10       label_seg [np.all(label==wall,axis=-1)] = 6
11       label_seg [np.all(label==polegroup,axis=-1)] = 7
12       label_seg [np.all(label==traffic_light,axis=-1)] = 8
13       label_seg [np.all(label==traffic_sign,axis=-1)] = 9
14       label_seg [np.all(label==terrain,axis=-1)] = 10
15       label_seg [np.all(label==vegetation,axis=-1)] = 11
16       label_seg [np.all(label==sky,axis=-1)] = 12
17       label_seg [np.all(label==person,axis=-1)] = 13
18       label_seg [np.all(label==rider,axis=-1)] = 14
19       label_seg [np.all(label==bicycle,axis=-1)] = 15
20       label_seg [np.all(label==bus,axis=-1)] = 16
21       label_seg [np.all(label==car,axis=-1)] = 17
22       label_seg [np.all(label==caravan,axis=-1)] = 18
23       label_seg [np.all(label==motorcycle,axis=-1)] = 19
24       label_seg [np.all(label==train,axis=-1)] = 20
25       label_seg [np.all(label==truck,axis=-1)] = 21
26       label_seg = label_seg[:,:,0]
27
28       return label_seg
```

Listing 4.3: Function for setting labels.

# 5

## EXPERIMENTAL RESULTS

This chapter presents the results of the methods described in the previous chapter. We want to discuss and analyze our experimental results for the scene understanding methods and the semantic segmentation models. First, we will discuss the dataset BDD100K and then evaluate each model with different approaches in terms of the loss function, accuracy, and mean Intersection over Union (mIoU).

## 5.1 Dataset used and Baseline

In this work, we have used the BDD100k dataset. This dataset was collected by The University of California, Berkeley. You can download the dataset through this link[1]. In order to evaluate the exciting progress of image recognition algorithms in autonomous driving, they construct the largest driving video dataset, BDD100K, with 100K videos and 10 tasks. Geographic, environmental, and weather diversity in the dataset makes it useful for training models that are less susceptible to being surprised by new conditions. As well as images with high resolution (720p) and frame rate (30 fps), a GPS/IMU recording are included in the dataset to preserve the driving trajectory. Generally, over 50K rides in New York, the San Francisco Bay Area, and other cities resulted in 100K driving videos (40 seconds each). A variety of scene types are included in the dataset, such as city streets, residential areas, and highways. Additionally, the videos were recorded at different times of the day and in different weather conditions [12]. Figure 5.1 illustrates of BDD100k dataset. In addition, we have used the CamVid (Cambridge-driving Labeled Video Database) dataset as an auxiliary dataset. Due to its small size, the CamVid dataset aimed us to Analyze the models more comfortably and to find out and fix the defects faster. The CamVid (Cambridge-driving Labeled Video Database) comprises five video sequences that were captured with a 960Œ720 resolution camera mounted on the dashboard of a car. A total of 701 frames were sampled (four at 1 frame per second (fps) and one at 15 fps). Annotating those stills manually with 32 classes resulted in the following: void, building, wall, tree, vegetation, fence, sidewalk, parking block, column/pole, traffic cone, bridge, sign, miscellaneous text, traffic light, sky, tunnel, archway, road, road shoulder, lane markings (driving), lane markings (non-driving), animal, pedestrian, child, cart luggage, bicyclist, motorcycle, car, SUV/pickup/truck, truck/bus, train, and other moving object [49].Figure 5.2 illustrates of CamVid dataset.

---

[1] https://bdd-data.berkeley.edu/

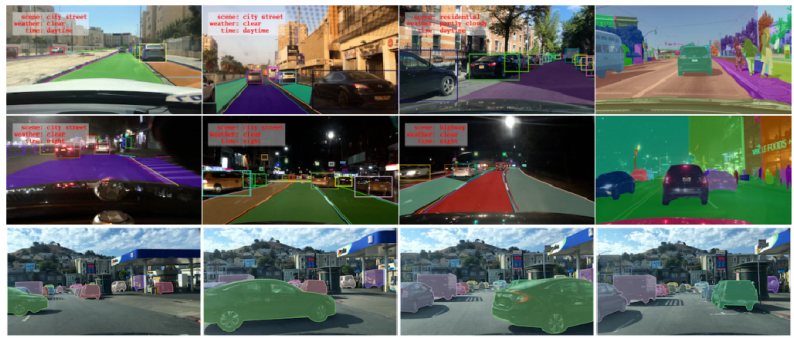

Figure 5.1: Overview of BDD100k dataset. Various driving videos are included in the BDD100k dataset under varying weather conditions, timings, and scene types [12].

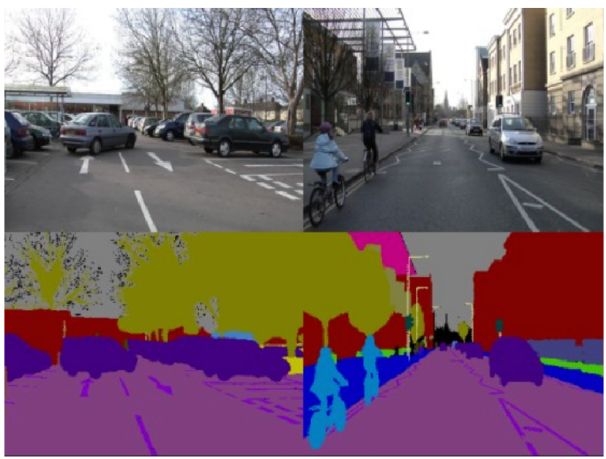

Figure 5.2: An example of CamVid dataset images.

### 5.1.1 Semantic Segmentation

According to [12], through joint training of semantic segmentation, detection, and lane marking/drivable area, the researchers fine-tune a base semantic segmentation model. Figure 5.3 illustrates of evaluation results for semantic segmentation. They used multi-task learning in this work. However, the means Intersection over Union (mIoU) for the semantic segmentation without other tasks is equal to 56.9. The researchers found that training with the additional 70K object detection dataset improved the overall mIoU from 56.9 to 58.3, with the improvement attributed mostly to the object classes in the object detection dataset. BDD100K's weather, scene, and time of day attributes are shown in Figure 5.4.

| Training Set | Road | Sidewalk | Building | Wall | Fence | Pole | Light | Sign | Vegetation | Terrain | Sky | Person | Rider | Car | Truck | Bus | Train | Motorcycle | Bicycle | mean IoU |
|---|---|---|---|---|---|---|---|---|---|---|---|---|---|---|---|---|---|---|---|---|
| Sem-Seg | 94.3 | 63.0 | 84.9 | 25.7 | 45.8 | 52.6 | 56.2 | 54.1 | **86.4** | 45.1 | 95.3 | 62.4 | 22.1 | 90.2 | 50.5 | 68.3 | 0 | 35.5 | **49.9** | 56.9 |
| Sem-Seg + Det | 94.3 | 62.5 | **85.2** | 24.5 | 41.1 | 51.5 | **63.1** | **57.9** | 86.2 | **47.4** | **95.5** | **64.6** | **28.1** | **90.8** | **52.9** | **70.7** | 0 | **43.4** | 48.9 | **58.3** |
| Sem-Seg + Lane + Driv | **94.8** | **65.8** | 84.1 | 22.6 | 40.2 | 49.3 | 51.9 | 49.7 | 85.8 | 46.2 | 95.3 | 60.8 | 7.1 | 89.9 | 47.8 | 66.9 | 0 | 27.5 | 27.5 | 53.3 |

Figure 5.3: Evaluation results for semantic segmentation [12].

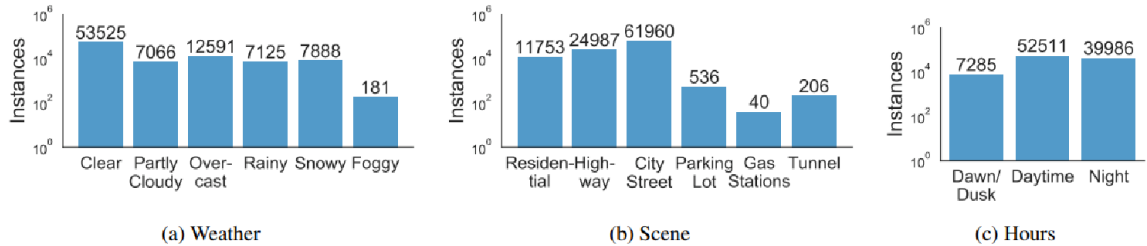

(a) Weather

(b) Scene

(c) Hours

Figure 5.4: The distribution of images according to weather conditions, scenes, and day hours [12].

According to the segmentation annotations, Figure 5.5 shows the distribution of the number of instances observed.

#### 5.1.1.1 Evaluation metrics for segmentation

#### IoU (Intersection over Union)

A common evaluation metric for semantic image segmentation is intersection-over-union. IoU or Jaccard Index are used to determine whether or not a prediction is correct.

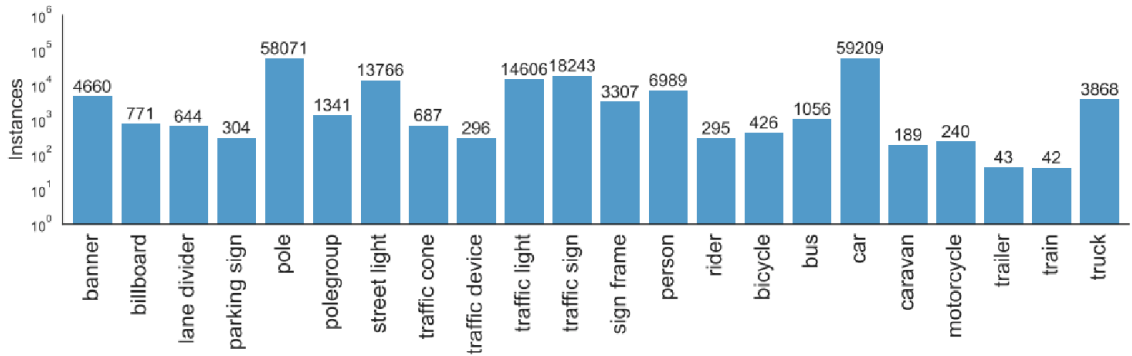

Figure 5.5: On the BDD100k dataset, the distribution of classes in semantic instance segmentation is shown [12].

Similarity of two sets $U$ and $V$. Iou is shown in Figure 5.6.

$$(5.1) \qquad iou\, Or\, Jaccard(U,V) = \frac{|U \cap V|}{|U \cup V|}$$

Or

$$(5.2) \qquad iou = true\text{-}positives/(true\text{-}positives + false\text{-}positives + false\text{-}negatives)$$

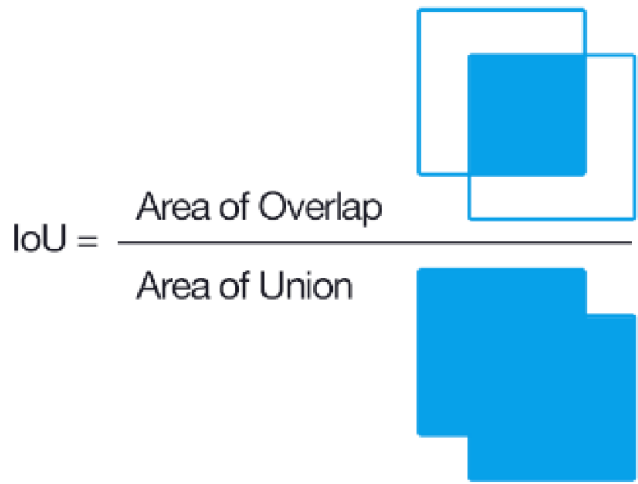

Figure 5.6: IoU (Intersection over Union).

**Accuracy**

Pixel accuracy is perhaps the easiest to measure. It showed the percent of pixels in your image that is classified correctly. It is important to remember that the accuracy criterion alone cannot be used for model comparison. The point is that high pixel accuracy does not necessarily imply superior segmentation capabilities.

$$(5.3) \qquad Accuracy = \frac{TP + TN}{TP + TN + FP + FN}$$

**Loss function**

Loss function is discussed in detail in section 2.3.

**F1 score**

In binary classification problems or problems with multiple binary labels or classes, the F1-score is used to assess the quality of the classification problem. The best value of the F1-score is 1 (perfect precision and recall), and the worst value is 0.

$$(5.4) \qquad F1 = \frac{2 * Precision * Recall}{Precision + Recall} = \frac{2 * TP}{2 * TP + FP + FN}$$

## 5.2 Training and Results

This section presents the results of training compounding models with different backbones on the BDD100k dataset. Several experiments were performed on models with different backbones during the work of this thesis. We proposed an innovative compounding of models with different backbones, and then we investigated and analyzed the obtained results. Tables 5.1 and 5.2 illustrates the performance evaluated on the training and validation throughout training. Also, the obtained results show that FPN with EfficientNet backbone (proposed method) achieves the highest accuracy score during the training and validation period. In addition, the lowest loss function is also for this approach (see table 5.3 and 5.4). The obtained results show that choosing the appropriate backbone has a great effect on the performance of the model for semantic segmentation. Better performance in semantic segmentation allows us to understand better the scene and the environment around the agent.

## 5.3  Scene Understanding Results

As discussed in 5.1.1, and 5.2, We used the FPN model with the EfficientNet backbone (proposed method) model for this stage of semantic segmentation. In the baseline paper [12], they train detection and instance segmentation at batch-level round-robin, using Mask R-CNN [50] and ResNet-50 as backbones. Table 5.1 demonstrates the performance of our proposed method on the segmentation task for the BDD100k dataset. Our proposed method improves the Sem-Seg mIoU from 56.9 to about 62.6. This is about a 5.7% increase in mIoU score. Also, our proposed method achieved a loss score of 0.092 on the BDD100k dataset. Some classes have more abundance. As a result, they include more pixels of the image. For this reason, the classes that have less abundance in the dataset get fewer scores. It is necessary to use loss functions along with weighting classes in order to solve this problem. By using this method, the classes are brought into balance.

The time required to run this program and obtain its results, according to the system specifications that we had used (figure 5.7), lasted about 130 hours.

| System Information | Value |
|---|---|
| CPU | Intel(R) Xeon(R) Silver 4210R CPU @ 2.40GHz |
| Memory | 128GiB System memory |
| GPU | 4*Tesla K80 − 1*Tesla V100S PCIe 32GB |
| System Architecture | x86_64 |
| OS | Pop!_OS 20.04 LTS |
| CPU MHz | 3140.524 |

Figure 5.7: Specifications of the system used to obtain the results

Table 5.1: The results of different implemented models have been reported here. Results show that FPN with EfficientNet backbone (proposed method) achieves the highest mean Intersection over Union (mIou).

| Model | Backbones | | Mean IoU |
|---|---|---|---|
| Unet | ResNet | resnet34 | 0.4255239 |
| | | resnet50 | 0.39371056 |
| | VGG | vgg16 | 0.36845574 |
| | | vgg19 | 0.3528282 |
| | DenseNet | densenet121 | 0.41398702 |
| | | densenet169 | 0.41919354 |
| | | densenet201 | 0.31873678 |
| | Inception | inceptionv3 | 0.397086 |
| | | inceptionresnetv2 | 0.36762936 |
| | MobileNet | mobilenet | 0.4263068 |
| | | mobilenetv2 | 0.31404237 |
| | EfficientNet | efficientnetb3 | 0.42914072 |
| | | efficientnetb4 | 0.43450747 |
| FPN | ResNet | resnet34 | 0.4287985 |
| | | resnet50 | 0.4228332 |
| | VGG | vgg16 | 0.37824065 |
| | | vgg19 | 0.32498728 |
| | DenseNet | densenet121 | 0.41112492 |
| | | densenet169 | 0.43958346 |
| | | densenet201 | 0.44255204 |
| | Inception | inceptionv3 | 0.4347405 |
| | | inceptionresnetv2 | 0.35110006 |
| | MobileNet | mobilenet | 0.42654864 |
| | | mobilenetv2 | 0.20107874 |
| | **EfficientNet** | **efficientnetb3** | 0.58567513 |
| | | **efficientnetb4** | **0.62574893** |

Table 5.2: The results of different implemented models have been reported here. Results show that FPN with EfficientNet backbone (proposed method) achieves the highest mean Intersection over Union (mIou).

| Model | Backbones | | Mean IoU |
|---|---|---|---|
| Linknet | ResNet | resnet34 | 0.3726975 |
| | | resnet50 | 0.3672478 |
| | VGG | vgg16 | 0.34751845 |
| | | vgg19 | 0.32641109 |
| | DenseNet | densenet121 | 0.39165006 |
| | | densenet169 | 0.42310557 |
| | | densenet201 | 0.4051628 |
| | Inception | inceptionv3 | 0.43081943 |
| | | inceptionresnetv2 | 0.4258381 |
| | MobileNet | mobilenet | 0.41240094 |
| | | mobilenetv2 | 0.2451828 |
| | EfficientNet | efficientnetb3 | 0.40906647 |
| | | efficientnetb4 | 0.4118332 |
| PSPNet | ResNet | resnet34 | 0.38286317 |
| | | resnet50 | 0.3893988 |
| | VGG | vgg16 | 0.34513064 |
| | | vgg19 | 0.33591891 |
| | DenseNet | densenet121 | 0.4005779 |
| | | densenet169 | 0.38262 |
| | | densenet201 | 0.31608737 |
| | Inception | inceptionv3 | 0.41286275 |
| | | inceptionresnetv2 | 0.40917594 |
| | MobileNet | mobilenet | 0.38976783 |
| | | mobilenetv2 | 0.22596618 |
| | EfficientNet | efficientnetb3 | 0.39239863 |
| | | efficientnetb4 | 0.4256185 |

Table 5.3: The results of different implemented models have been reported here. Results show that FPN with EfficientNet backbone (proposed method) achieves the highest accuracy score during the training and validation period.

| Model | Backbones | | Val_Accuracy | Loss |
|---|---|---|---|---|
| Unet | ResNet | resnet34 | 0.8430 | 0.1438 |
| | | resnet50 | 0.8183 | 0.1704 |
| | VGG | vgg16 | 0.7918 | 0.2538 |
| | | vgg19 | 0.7857 | 0.2696 |
| | DenseNet | densenet121 | 0.8433 | 0.1925 |
| | | densenet169 | 0.8482 | 0.1733 |
| | | densenet201 | 0.7216 | 0.2185 |
| | Inception | inceptionv3 | 0.8377 | 0.1731 |
| | | inceptionresnetv2 | 0.7832 | 0.4520 |
| | MobileNet | mobilenet | 0.8462 | 0.1343 |
| | | mobilenetv2 | 0.7020 | 0.2087 |
| | EfficientNet | efficientnetb3 | 0.8487 | 0.1998 |
| | | efficientnetb4 | 0.8580 | 0.1517 |
| FPN | ResNet | resnet34 | 0.8479 | 0.0836 |
| | | resnet50 | 0.8385 | 0.1025 |
| | VGG | vgg16 | 0.8104 | 0.2207 |
| | | vgg19 | 0.7745 | 0.3638 |
| | DenseNet | densenet121 | 0.8244 | 0.1376 |
| | | densenet169 | 0.8560 | 0.1023 |
| | | densenet201 | 0.8533 | 0.0989 |
| | Inception | inceptionv3 | 0.8498 | 0.1456 |
| | | inceptionresnetv2 | 0.7795 | 0.5082 |
| | MobileNet | mobilenet | 0.8480 | 0.1031 |
| | | mobilenetv2 | 0.4382 | 0.2073 |
| | **EfficientNet** | **efficientnetb3** | 0.8704 | 0.1118 |
| | | **efficientnetb4** | **0.8976** | **0.0919** |

Table 5.4: The results of different implemented models have been reported here. Results show that FPN with EfficientNet backbone (proposed method) achieves the highest accuracy score during the training and validation period.

| Model | Backbones | | Val_Accuracy | Loss |
|---|---|---|---|---|
| Linknet | ResNet | resnet34 | 0.7830 | 0.1015 |
| | | resnet50 | 0.8112 | 0.1988 |
| | VGG | vgg16 | 0.7881 | 0.4852 |
| | | vgg19 | 0.7732 | 0.4289 |
| | DenseNet | densenet121 | 0.8107 | 0.3176 |
| | | densenet169 | 0.8475 | 0.2090 |
| | | densenet201 | 0.8376 | 0.2144 |
| | Inception | inceptionv3 | 0.8393 | 0.2003 |
| | | inceptionresnetv2 | 0.8499 | 0.1658 |
| | MobileNet | mobilenet | 0.8430 | 0.1594 |
| | | mobilenetv2 | 0.5382 | 0.3150 |
| | EfficientNet | efficientnetb3 | 0.8284 | 0.1873 |
| | | efficientnetb4 | 0.8384 | 0.1658 |
| PSPNet | ResNet | resnet34 | 0.8220 | 0.1746 |
| | | resnet50 | 0.8257 | 0.1789 |
| | VGG | vgg16 | 0.7902 | 0.3510 |
| | | vgg19 | 0.7682 | 0.3540 |
| | DenseNet | densenet121 | 0.8327 | 0.1789 |
| | | densenet169 | 0.8226 | 0.2351 |
| | | densenet201 | 0.7105 | 0.4244 |
| | Inception | inceptionv3 | 0.8264 | 0.1715 |
| | | inceptionresnetv2 | 0.8322 | 0.1537 |
| | MobileNet | mobilenet | 0.8084 | 0.2043 |
| | | mobilenetv2 | 0.5681 | 0.3632 |
| | EfficientNet | efficientnetb3 | 0.8243 | 0.2172 |
| | | efficientnetb4 | 0.8392 | 0.1979 |

Table 5.5: Hyperparameters for our proposed compound model.

| Hyperparameter | Value |
|---|---|
| Activation | softmax |
| Learning rate | 0.0001 |
| Optimizer | Adam |
| IOUScore (threshold) | 0.5 |
| FScore (threshold) | 0.5 |
| Loss | categorical_crossentropy |
| Batch_size | 8 |
| Epochs minimum | 50 |

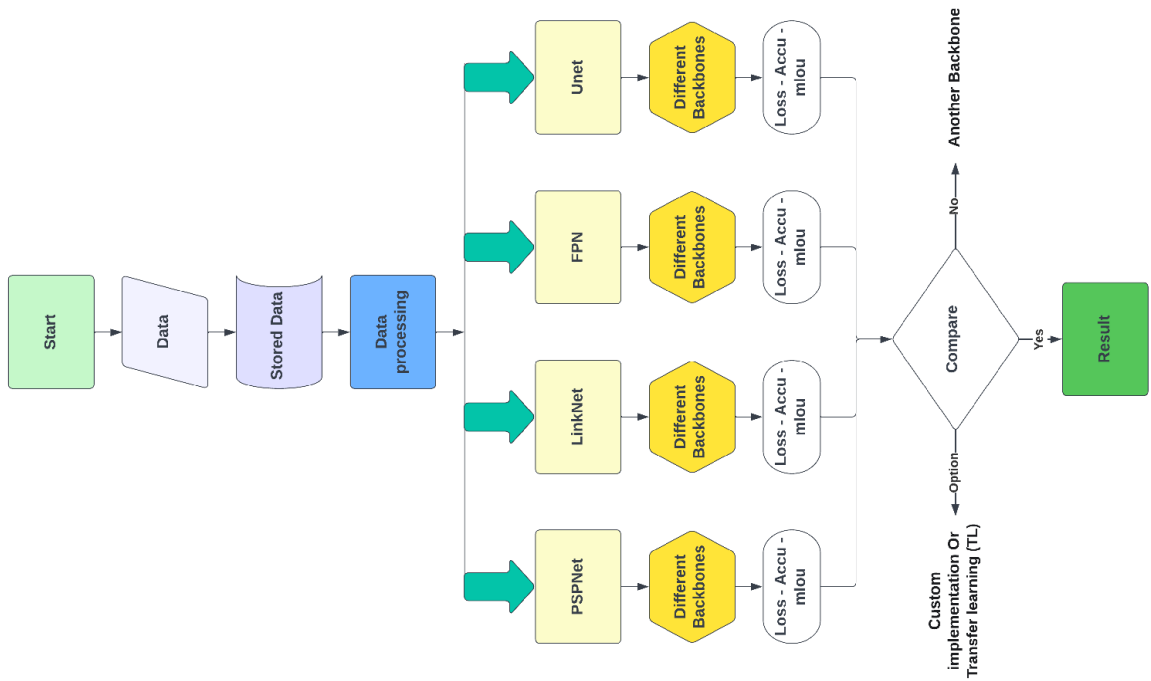

Figure 5.8: Progress of work

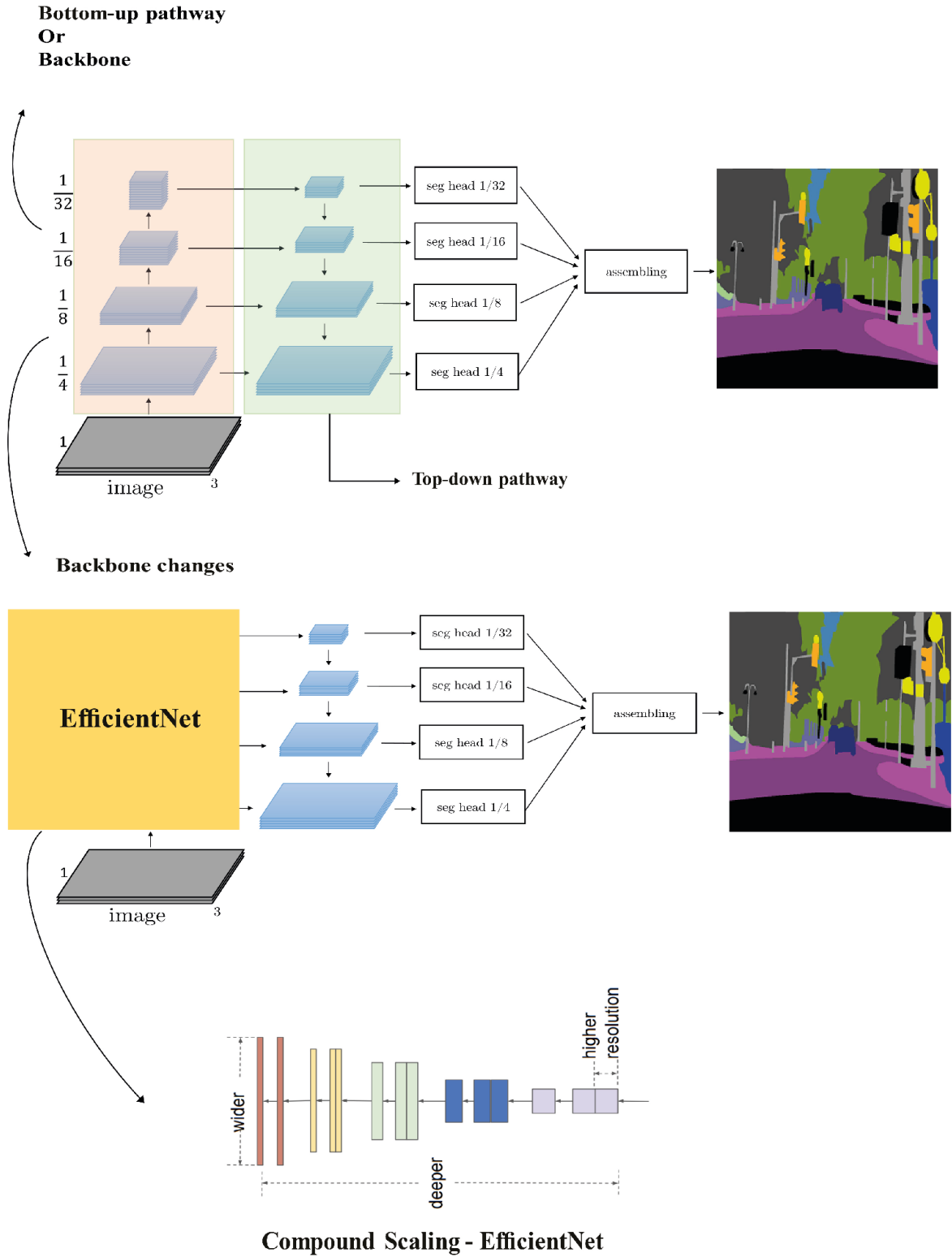

Figure 5.9: The architecture of the proposed approach is in this thesis

## CONCLUSION AND FUTURE WORK

For self-driving cars are being allowed to pass through the city without any supervision and control, it is necessary to have a proper understanding and analysis of the environment. As much as the car (agent) can better understand the objects and pedestrians in the scene and on the roads, safety will increase similarly proportioned. correct and timely recognition of road objects enables faster and more rational decisions to be made. Semantic segmentation is one of the main subsections of scene understanding.

In this work, we propose several efficient models to investigate scene understanding through semantic segmentation. We use the BDD100k dataset to investigate these models. In this thesis, we proposed an innovative compounding of models with different backbones, and then we investigated and analyzed the obtained results. The obtained results show that choosing the appropriate backbone has a great effect on the performance of the model for semantic segmentation. Better performance in semantic segmentation allows us to understand better the scene and the environment around the agent. Our best compound model for semantic segmentation on BDD100k datasets was the Feature Pyramid Networks (FPN) model with the EfficientNet backbone that achieved the highest level of performance in our task. We hope that the review and analysis of the obtained results would be helping to improve the performance of models in the field of semantic segmentation. Further, these results could serve as a helpful perspective for future research in other fields such as medical images, cancer diagnosis, satellite images, robotic navigation, localization, traffic control systems, etc.

The following open challenges could aim as useful directions for future work in the field of scene understanding and semantic segmentation:

**1.** Build a new snowy weather conditions dataset for Autonomous Vehicles with the following tools:

- Labkit: https://imagej.net/plugins/labkit/

- QuPath: https://qupath.github.io/

- Label Studio: https://labelstud.io/

**2.** Evaluate our proposed compound model on only the snowy condition dataset.

**3.** Can our proposed compound model be effective in multi-task learning?

**4.** Can our proposed approach be effective in improving the diagnosis of cancer cells and their segmentation?

**5.** Can our proposed approach be effective in improving the semantic segmentation of aerial satellite imagery?

To investigate our proposed model on the data that defined and labeled snow conditions as a class, you can take help from datasets Mapillary Vistas Dataset and Canadian Adverse Driving Conditions Dataset.

- **Mapillary Vistas Dataset:** This is a novel, large-scale dataset of street-level images, which contains 25000 high-resolution images annotated with 66 object categories and 37 instance-specific labels. By using polygons to delineate individual objects, annotation is performed densely and fine-grained. Images from around the world are captured under varying weather, season, and daytime conditions in their dataset, which is 5 times larger than Cityscapes' total amount of fine annotations. It is important to note that images are captured from a variety of different imaging devices (mobile phones, tablets, action cameras, professional capturing rigs) and by different photographers with different experience levels. Thus, the dataset is designed and compiled in such a way that it covers diversity, richness, and geography. In addition to semantic image segmentation, instance-specific image segmentation is defined as a default benchmark task, which aims to significantly advance the development of current road-scene understanding methods [51].

- **Canadian Adverse Driving Conditions Dataset:** This paper [52] presents the CADC dataset, which contains lidar and images collected during winter driving conditions within the Region of Waterloo. In challenging winter weather conditions, researchers will be able to test their object detection, localization, and mapping techniques using this dataset. A 2D annotation containing truncation and occlusion values will be released for each image in their future plans. Using this dataset, they will also develop a benchmark for 3D object detection [52].

Also, The Multimodal Brain Tumor Image Segmentation Benchmark (BRATS) [53], The Liver Tumor Segmentation Benchmark (LiTS) [54], CSAW-S (dataset of mammography images) [55], etc datasets can be used to investigate our proposed model in the field of medicine and cancer and tumor diagnosis.

Further research into satellite images can be conducted using the iSAID dataset. This paper [56] presents the first benchmark dataset for combining object detection on an instance level with pixel-level segmentation on aerial imagery. Aerial images present unique challenges when segmenting instances, for example, the high number of instances per image, the large-scale variations of objects, and the abundance of small objects. There are 655,451 object instances for 15 categories across 2,806 high-resolution images in their large-scale and densely annotated Instance Segmentation in Aerial Images Dataset (iSAID). For detailed scene analysis, such precise per-pixel annotations are essential for accurate localization. There are 15 times more categories of objects and 5 times more instances in iSAID than are present in existing small-scale aerial image-based instance segmentation datasets [56].